\documentclass[journal,twoside]{IEEEtran}

\usepackage{amsmath,amssymb,amsfonts}
 \usepackage{multirow} 
\usepackage{cite}
\usepackage{array}
\usepackage{graphicx}
\usepackage{booktabs}
\usepackage{svg}
\usepackage{textcomp}
\usepackage{url}
\usepackage{stfloats}
\usepackage{hyperref}
\usepackage[caption=false,font=normalsize,labelfont=sf,textfont=sf]{subfig}
\usepackage[table]{xcolor}
\usepackage{balance}
\usepackage{threeparttable}
\usepackage{soul}
\soulregister{\cite}7
\soulregister{\ref}7
\usepackage[ruled,vlined]{algorithm2e}

\definecolor{ieeegray}{gray}{0.35}

\def\SFM1#1{{\bf [Sifat:} {\it\color{purple} {#1}}{\bf ]}}

\definecolor{rev}{RGB}{0,0,0} 
\def\rev#1{{\color{rev} #1}}
\def\rvn#1{{\color{rev} #1}}

\newcommand{\berlinsize}{\fontsize{6}{7}\selectfont}

\begin{document}
\bstctlcite{IEEEexample:BSTcontrol}
 
\title{AGSA-Net: Abundance-Guided Self-Attention Network for Spectral
Unmixing-Aware Hyperspectral Remote Sensing Image Classification}

\author{
Nafisa Anjum,
Satavisa Dey Borno,
Ananna Saha,
Mir Faiyaz Hossain, \\
Sifat Momen,
Nabeel Mohammed,
and Shafin Rahman
\thanks{Nafisa Anjum, Satavisa Dey Borno, Ananna Saha, and Mir Faiyaz Hossain are graduates of North South University, Dhaka, Bangladesh
(e-mail: nafisa.anjum12@northsouth.edu; satavisa.borno@northsouth.edu; ananna.saha@northsouth.edu; mir.hossain01@northsouth.edu).}%
\thanks{Sifat Momen and Nabeel Mohammed are Professors with the Department of Electrical and Computer Engineering, North South University, Dhaka, Bangladesh
(e-mail: sifat.momen@northsouth.edu; nabeel.mohammed@northsouth.edu).}%
\thanks{Shafin Rahman is an Associate Professor with the Department of Electrical and Computer Engineering, North South University, Dhaka, Bangladesh
(e-mail: shafin.rahman@northsouth.edu).}
}

\maketitle

\begin{abstract}
\label{sec: abstract}
Hyperspectral image (HSI) classification plays a vital role in remote sensing applications, including agriculture, environmental monitoring, and urban analysis. However, its performance remains challenged by high spectral redundancy, noise sensitivity, and the difficulty of jointly modeling local material composition and long-range spectral dependencies. To address this, we propose AGSA-Net, an abundance-guided self-attention network that explicitly integrates spectral unmixing priors into the classification process. AGSA-Net first estimates physically meaningful subpixel abundance maps subject to non-negativity and sum-to-one constraints, regularized by hybrid linear-nonlinear reconstruction decoder. The learned abundances are then used to construct an abundance affinity prior that guides a spectral transformer to emphasize class-discriminative interactions, and the resulting transformer features are fused with compact abundance descriptors for final prediction; in contrast to existing approaches that use abundance as auxiliary or concatenated features. \rev{ Experiments on Indian Pines, Augsburg, and Berlin demonstrate the benefit of incorporating abundance-guided contextual modeling, particularly in heterogeneous urban scenes.}
The source code and trained models are available at: \url{https://github.com/nnuvi/AGSA-Net}

\end{abstract}


\section{Introduction}
\label{sec:intro}


\IEEEPARstart{R}{emote} \rev{sensing plays a critical role in  Earth observation, supporting applications such as land mapping, environmental monitoring, disaster assessment, and precision agriculture \cite{rsLandMapping, envmotitor, soil, Crop, bioucas2013hyperspectral}. Hyperspectral imaging (HSI) is particularly valuable because it captures hundreds of narrow, contiguous spectral bands, enabling finer material discrimination than multispectral or RGB imagery \cite{bioucas2013hyperspectral, iMAGEspec}. The core task of HSI classification is to assign semantic labels to pixels based on their spectral-spatial characteristics to generate accurate land-cover maps. However, this task remains challenging due to high spectral dimensionality, strong band correlations, spatial heterogeneity, atmospheric scattering, illumination variation, sensor noise, limited labeled samples, and mixed pixels \cite{EndmemberVar, Interference}. These factors increase inter-class similarity and intra-class variability, making robust feature extraction and generalization difficult.}

\rev{
Deep learning has significantly advanced HSI classification through hierarchical spectral-spatial feature learning. Representative methods include 3D CNNs \cite{3DCNN, DLCNN}, recurrent networks \cite{RNN}, graph convolutional networks \cite{GCN, CNNGNN, GANN}, and transformer-based models for long-range dependency modeling \cite{19WAVE, SpectraForner}. Despite their success, many approaches rely heavily on labeled data and compute similarity mainly from raw spectral or learned feature representations. This can be unreliable in hyperspectral scenes affected by spectral redundancy and mixed pixels, where spectrally similar pixels may have different underlying material compositions. To alleviate these issues, unmixing-inspired autoencoders have been introduced to model sub-pixel material composition \cite{Palsson2020CAE, CYCU2021, Wang2019NonlinearAE}. Their estimated abundances are commonly used as auxiliary features for classification \cite{MaskedAuto, DCNET}.
}

\rvn {
In existing approaches, abundance information is typically treated as a preprocessing step or simply concatenated with learned features, without being explicitly integrated into contextual reasoning mechanisms, limiting the exploitation of material composition for discriminating spectrally similar classes. Existing transformer-based HSI classifiers compute attention primarily from spectral or learned feature similarity. However, hyperspectral scenes frequently contain mixed pixels and strong spectral redundancy, where spectrally similar pixels may correspond to different underlying material compositions. In such cases, conventional self-attention may form unreliable contextual relationships by emphasizing spectral proximity alone. This limitation becomes particularly problematic in urban and boundary regions containing heterogeneous material mixtures. These observations suggest that material composition provides information beyond spectral similarity alone. Since abundance estimates describe the proportion of constituent materials within a pixel, abundance affinity can better reflect material-level relationships between pixels, particularly in mixed-pixel regions where spectral similarity alone may be ambiguous. Despite this potential, material composition is not explicitly incorporated into the attention formation process of most existing transformer-based HSI models.
}
 
\begin{figure*}[!t]
\centering
\begin{minipage}[c]{0.49\textwidth}
    \centering
    \includegraphics[width=\linewidth]{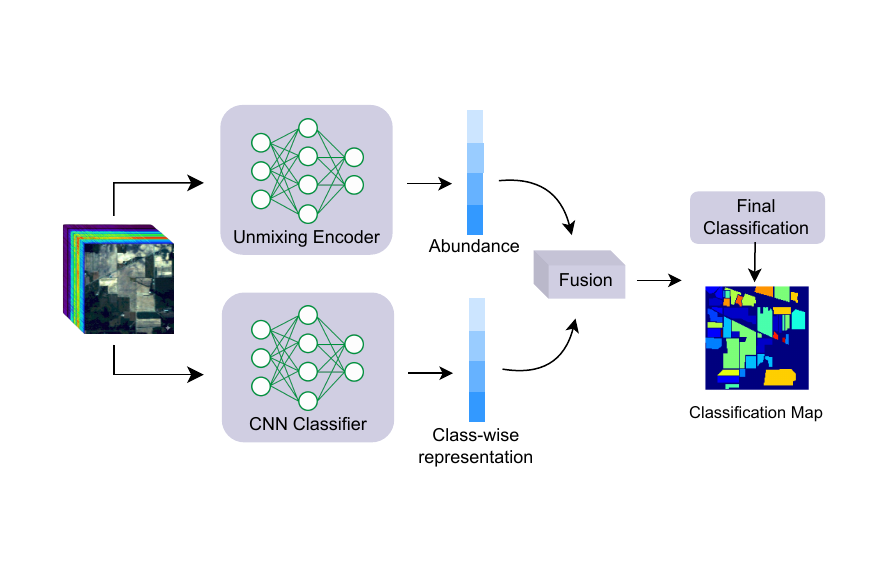}
    \small (a)
    \par\vspace{3pt}
\end{minipage}\hfill
\begin{minipage}[c]{0.51\textwidth}
    \centering
    \includegraphics[width=\linewidth]{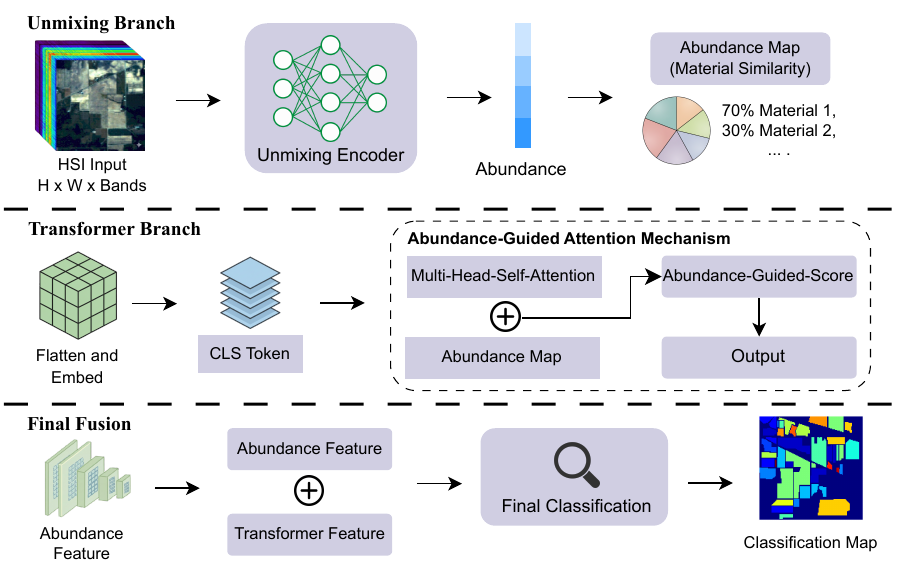}
    \small (b)
    \par\vspace{3pt}
\end{minipage}
\caption{Overview of the proposed AGSA-Net for HSI classification. The framework comprises an unmixing branch and a transformer branch. (a) In contrast to prior approaches that treat abundance maps only as auxiliary inputs or concatenated features, (b) AGSA-Net converts the estimated abundances into a pixel-wise material affinity matrix and injects it directly into the multi-head self-attention as an additive bias, thereby explicitly governing attention formation. This design enforces contextual interactions driven by sub-pixel material composition rather than spectral similarity alone. Final predictions are obtained by fusing abundance-aware transformer features with compact abundance representations.
}
\label{fig:fig_1}
\end{figure*}

\rvn {
Building upon this insight, we propose AGSA-Net (Abundance-Guided Self-Attention Network), a dual-branch framework that couples spectral unmixing with transformer-based classification as illustrated in Fig.~\ref{fig:fig_1}. Following DSNet (Dual-Branch Subpixel-Guided Network) [21], AGSA-Net adopts the same unmixing autoencoder, abundance estimation strategy, and SAD-based supervision to obtain physically meaningful abundance maps. However, unlike DSNet, which uses abundances only for downstream feature fusion, AGSA-Net converts abundance representations into a pixel-wise material-affinity matrix and injects it directly into multi-head self-attention as an additive bias. This design enables the transformer to model contextual relationships based on sub-pixel material composition rather than spectral similarity alone, improving discrimination between spectrally similar but compositionally distinct classes.
}

\rvn {
We evaluate AGSA-Net on benchmark datasets using Overall Accuracy (OA), Average Accuracy (AA), and the Kappa coefficient. Experimental results demonstrate competitive performance across diverse agricultural and urban scenes.
}

The main contributions of this paper are summarized as follows:
\begin{itemize}

    \item \textbf{Dual-branch integration of spectral unmixing and classification:} \textcolor{black}{To address the challenges of mixed pixels, spectral ambiguity, and unreliable spectral similarity}, we propose \textbf{AGSA-Net}, where spectral unmixing from the autoencoder is directly coupled with transformer-based classification instead of being used only as an auxiliary input feature. \textcolor{black}{  Spectral redundancy and long-range dependency modeling challenges are handled by the spectral transformer backbone,    enabling global contextual interactions.}
    
    \item \textbf{Abundance-guided attention and feature fusion:}
    We inject abundance affinity as an additive bias in self-attention and jointly fuse abundance features with transformer representations to enable subpixel-aware spectral-spatial modeling, \textcolor{black}{ to overcome limitations in contextual reasoning and long-range dependency modeling under spectral redundancy.}
    \item \textbf{Validated performance across benchmarks:}
    We conduct extensive experiments on the Indian Pines, Augsburg, and Berlin datasets, demonstrating consistent improvements over several recent state-of-the-art methods. 
\end{itemize}

The remainder of this paper is organized as follows.
Section~\ref{sec:related} introduces related works.
Section~\ref{sec:method} presents the methodology of AGSA-Net.
Experiment results and analyses are shown in Section~\ref{sec:experiments}.
Finally, Section~\ref{sec:conclusion} concludes the paper.

\section{Related Works}
\label{sec:related}
\subsection{\textcolor{black}{Machine Learning} Approaches in HSI Remote Scene Classification}
Early research in hyperspectral image (HSI) classification for remote sensing emphasized machine learning methods for spectral feature extraction. Support Vector Machines (SVMs) offered strong generalization with limited data \cite{SVM}, while Extended Morphological Profiles enhanced spatial descriptors for land-cover discrimination \cite{EMP}. Subsequent techniques included sparse representation \cite{SRB}, Local Binary Patterns with Extreme Learning Machines \cite{ELM}, and other handcrafted features balancing efficiency and accuracy. However, these approaches struggled with complex spectral-spatial interactions, prompting a transition to deep learning.

\subsection{Deep Learning Approaches}
\noindent\subsubsection*{CNN-based models}
The introduction of Convolutional Neural Networks (CNNs) marked a significant advancement in remote sensing by enabling hierarchical spectral-spatial feature learning \cite{DLCNN, 3DCNN }. Recurrent neural networks further exploited sequential spectral relationships \cite{RNN}, including single GRU-based spectral-spatial classifiers for efficient processing \cite{gru}, while unified deep spectral-spatial architectures improved contextual modeling \cite{SSUN}. Despite these gains, CNN-based models often required large amounts of training data and had limited capacity to capture long-range dependencies.
\noindent\subsubsection*{Graph-based and Hybrid Models}
To address these shortcomings, graph-based learning emerged as a powerful alternative. Graph Convolutional Networks (GCNs) modeled non-Euclidean spatial structures \cite{GCN}, and later works introduced hybrid CNN-GNN fusion \cite{CNNGNN}, multi-scale graph attention mechanisms \cite{GANN}, and feature-fusion graph architectures \cite{WFCG}, significantly improving spatial consistency and robustness. These methods demonstrated strong performance but often at the cost of increased computational burden.
\noindent\subsubsection*{Transformer-based Models}
Transformer architectures introduced self-attention for global dependencies \cite{ViT}, while SpectralFormer \cite{SpectraForner} and WaveFormer \cite{19WAVE} focused on spectral long-range modeling. Masked autoencoders with contrastive learning \cite{MaskedAuto}, dual-branch transformers with LiDAR \cite{LiDAR}, and spectral-spatial tokenization \cite{SSFTT} expanded multimodal flexibility. Subpixel-guided networks \cite{DCNET}, capsule-based adversarial models \cite{DCCGAN}, and morphological convolutions \cite{MorphCONVO} provided complementary advances.

\subsection{Emerging Trends and Hybrid Techniques}
\label{subsec:lit@emerging_trends}
\textcolor{black}{Recent studies have increasingly explored transformer-based architectures for hyperspectral image (HSI) classification, demonstrating improved capability in modeling long-range spectral–spatial dependencies compared to conventional CNNs \cite{GraphGST, PSFormer}. Several works further enhance this paradigm by integrating hybrid frameworks that combine convolutional feature extraction with transformer attention to balance local detail preservation and global context modeling \cite{HyperMamba, CACFTNet}.}
\rvn{
To address the challenges of limited labeled samples, recent approaches have introduced self-supervised learning, contrastive strategies, and lightweight attention mechanisms to improve generalization under data-scarce conditions \cite{MCTGCL, GTCFN}. In parallel, graph-based and affinity-guided methods have been proposed to capture relational structures between pixels, enabling more effective modeling of spatial and spectral correlations \cite{MHS-Mamba, S2Mamba}. Another line of research incorporates spectral unmixing or abundance estimation into deep learning frameworks, leveraging sub-pixel material information to improve class separability in mixed-pixel scenarios \cite{MambaHSI+}. Spectral-spatial contrastive models \cite{CL}, cross-modality frameworks \cite{CMCL}, and self-supervised approaches \cite{cssl} further improve representation learning and generalization under limited supervision.
} 

Multifeature hybrids with KELM \cite{KELM}, capsule GANs \cite{DCCGAN}, residual attention networks \cite{RSSAN}, and fusion strategies \cite{MFFCG} bridge classical efficiency and deep expressivity. Research by Ghamisi et al. \cite{Advances} and Paoletti et al. \cite{DL} overview classical to deep paradigms, highlighting trends in spectral-spatial modeling, feature fusion, and scarce-supervision robustness.

\subsection{Limitations of Existing Works and Our Approach}
\label{subsec:lit@limit_existing_work_our_approach}
Despite significant progress in hyperspectral remote sensing scene classification, several limitations remain. \textcolor{black}{CNN-based models primarily capture local spectral-spatial patterns and struggle to model long-range contextual dependencies.} 
\rev{Transformer-based methods improve global context modeling but typically rely on feature similarity alone, which can be unreliable in hyperspectral imagery affected by mixed pixels and spectral redundancy. Unmixing-assisted classification methods estimate material abundances but often treat them as auxiliary inputs rather than integrating them into contextual reasoning mechanisms \cite{Palsson2020CAE}}

Deep models, such as CNNs, GNNs, and transformers, typically require large, labeled datasets, which are impractical in annotation-scarce remote sensing scenarios \cite{DL,cssl}. Although attention- and graph-based methods improve spectral-spatial representations \cite{GCN, CNNGNN, RSSAN}, they often struggle with spectral redundancy, sensor noise, class imbalance, and the need for long-range contextual modeling \cite{bioucas2013hyperspectral}. \rev{The proposed AGSA-Net integrates abundance-based material affinity as a guidance signal within the self-attention mechanism. By leveraging sub-pixel composition information together with spectral features, the framework seeks to enhance contextual representation learning in hyperspectral imagery.}

\section{Methodology}
\label{sec:method}
\subsection{Problem Formulation}
\label{subsec:pbf}
We consider a hyperspectral remote sensing image (HSI) 
$\mathbf{X} \in \mathbb{R}^{H \times W \times C}$ with spatial dimensions $H \times W$ and $C$ spectral bands, acquired by an airborne or satellite sensor for land-cover observation. 
Each pixel at spatial location $(i,j)$ is represented by a spectral vector $\mathbf{x}_{i,j} \in \mathbb{R}^{C}$, corresponding to surface reflectance measurements of terrestrial materials, and is associated with a ground-truth class label 
$y_{i,j} \in \mathcal{Y}$, where $\mathcal{Y} = \{y_1, \dots, y_U\}$ denotes the set of $U$ unique classes in the dataset representing land-cover or land-use categories. 
To incorporate local spatial context, we extract patches from the HSI. Specifically, let $\mathbf{P}_{i,j} \in \mathbb{R}^{p \times p \times C}$ denote a square patch of size $p \times p$ centered at pixel $(i,j)$. 
Each patch contains the spectral vectors of all pixels within the local spatial neighborhood, and its label is defined as the class of the central pixel, i.e., $y_{i,j}$. 
Given a patch $\mathbf{P}_{i,j}$, our objective is to predict the corresponding class label 
$\hat{y}_{i,j} \in \mathcal{Y}$. 
To this end, we train a dual-branch framework that estimates the posterior class distribution 
$p(y \mid \mathbf{P}_{i,j})$ for all $y \in \mathcal{Y}$ and assigns each patch to the class with the maximum probability.

Recent advances in hyperspectral remote scene classification have shown that modeling sub-pixel material composition can significantly improve discriminative representation learning. Prior works \cite{DCNET,DAEN2019,Wang2019NonlinearAE,Palsson2020CAE,CYCU2021} demonstrated that an unmixing-inspired autoencoder branch can extract physically meaningful abundance information, providing a stable and interpretable characterization of underlying material distributions. \rev{Among these approaches, DSNet \cite{DCNET} employs an autoencoder-based unmixing framework with SAD-based reconstruction supervision to obtain abundance representations from the unmixing branch. Building upon this framework, the proposed AGSA-Net adopts the same unmixing framework, SAD-based reconstruction supervision, and abundance feature fusion strategy, while additionally transforming the estimated abundances into a material-affinity prior that is directly incorporated into the self-attention computation. Consequently, contextual feature interactions are guided by sub-pixel material composition rather than relying solely on spectral similarity.} Using the adopted unmixing branch \cite{DCNET}, abundance maps \rev{are} estimated within each patch $\mathbf{P}_{i,j}$. However, this branch cannot encode spectral-spatial contextual relationships across pixels within the patch. As a result, pixels with similar spectral signatures but different material compositions still challenge discriminative models. \rev{To address this limitation,} we inject abundance-derived material affinity directly into the attention mechanism of a spectral transformer. This design enables contextual reasoning guided by physically meaningful material similarity, allowing the model to explicitly differentiate spectrally similar classes with distinct sub-pixel compositions. By combining an unmixing autoencoder with an abundance-guided spectral transformer, we obtain more discriminative hyperspectral representations.

\subsection{Model Overview}
\begin{figure*}[!t]
\centering
\includegraphics[width=\textwidth]{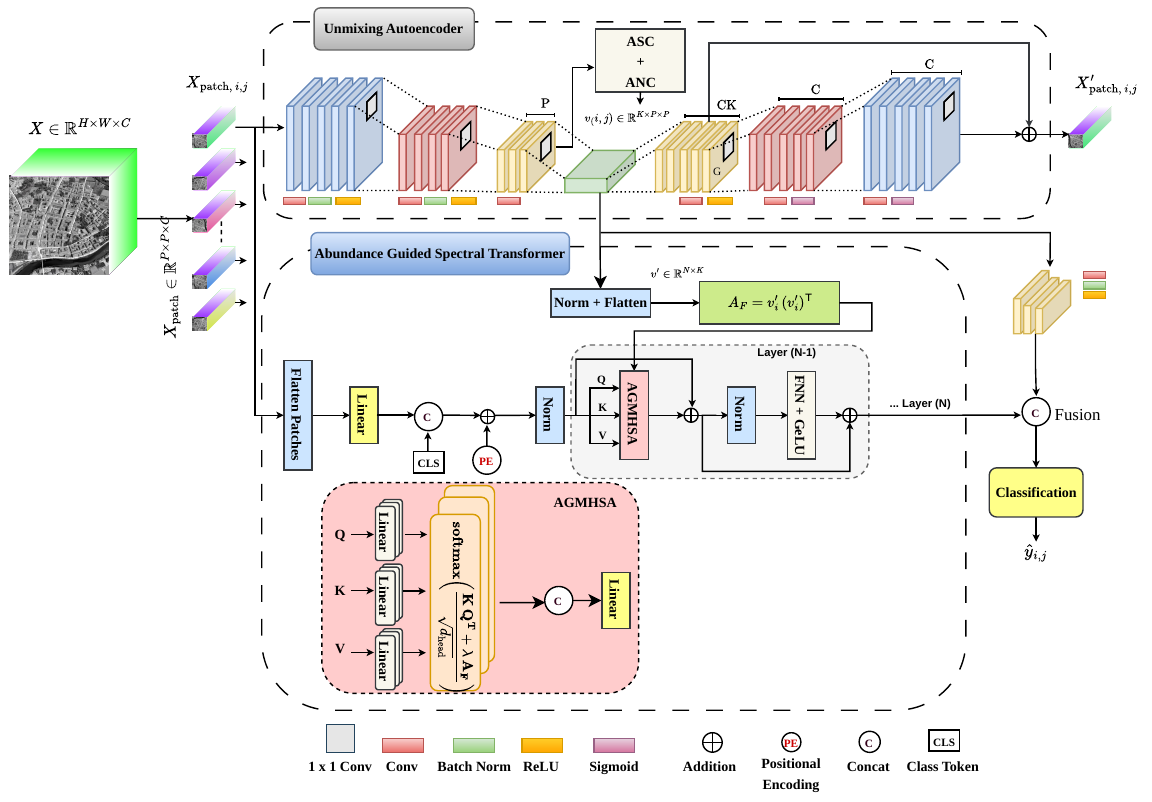}
\caption{Proposed AGSA-Net architecture. Given an input patch $\mathbf{P}_{i,j}\in\mathbb{R}^{p\times p\times C}$, the encoder in the first branch estimates pixel-wise abundances $\mathbf{v}_{i,j}\in\mathbb{R}^{K\times p\times p}$, which are flattened and normalized to compute the abundance affinity matrix. The decoder then reconstructs the patch using both linear and nonlinear mixing modules. In parallel, the spectral transformer tokenizes the same patch, projects the tokens into a latent space, and applies multi-head self-attention guided by the padded affinity matrix $\mathbf{AF}$ through an additive bias term. The classification token from the final transformer layer is fused with the abundance-derived features to produce the final logits.}
\label{fig:architecture}
\end{figure*}

Hyperspectral data consist of hundreds of narrow bands capturing fine-grained material signatures that lie beyond the visible spectrum. For robust pixel-level classification, it is essential to leverage rich spectral information jointly with local spatial structure, particularly in remote sensing scenes affected by mixed pixel spectra. To this end, we propose \textbf{AGSA-Net}, a dual-branch architecture illustrated in Fig.~\ref{fig:architecture}. The \textit{first branch} performs pixel-wise spectral unmixing using an autoencoder module, producing abundance maps that provide physically interpretable estimates of sub-pixel material composition. Unlike \cite{DCNET} and prior autoencoder-based unmixing approaches that only use abundances as auxiliary features, AGSA-Net transforms these abundances into pixel-pixel material affinity. This design couples unmixing with contextual reasoning, enabling the network to focus on pixels that share similar material composition and thereby enhancing discrimination between spectrally similar but compositionally distinct classes.
Given a patch $\mathbf{P}_{i,j}$, the encoder maps each pixel spectrum to a $K$-dimensional abundance vector, producing an abundance cube $\mathbf{v}_{i,j} \in \mathbb{R}^{K \times p \times p}$ whose entries represent normalized sub-pixel proportions of $K$ endmembers. These abundances are then used to calculate a pixel-pixel affinity matrix, which reflects material composition similarity.
The \textit{second branch} is a transformer-based classifier that takes spectral tokens extracted from patches  $\mathbf{P}_{i,j}$ and modulates its attention weights using the abundance-derived affinity structure. Specifically, the affinity matrix guides the attention operator, enabling the classifier to focus on pixels with similar underlying material mixtures rather than relying solely on raw spectral proximity. By leveraging abundance-guided multi-head self-attention, the model produces more discriminative contextual representations, leading to improved accuracy in land-cover classification of hyperspectral scenes.

\subsection{Unmixing Autoencoder}
\label{subsec:unmixenc}
The first branch of our proposed model adopts the unmixing autoencoder architecture from \cite{DCNET} to estimate physically meaningful abundance maps. Each pixel spectrum is mapped to a $K$-dimensional abundance vector that represents fractional contributions of latent endmembers. 
{\color{black}In this work, the number of endmembers $K$ is set equal to the number of semantic classes in the dataset, which provides a stable and interpretable abundance representation for the classification task.}

The encoder consists of $1\times1$ convolutions operating along the spectral dimension, ensuring pixel-wise abundance estimation while preserving spatial structure.
Let $\mathbf{v}_{i,j} \in \mathbb{R}^{K \times p \times p}$ denote the abundance tensor estimated from patch $\mathbf{P}_{i,j}$:
\begin{equation}
\mathbf{v}_{i,j} = f_{\theta_E}(\mathbf{P}_{i,j}).
\end{equation}

Non-negativity and sum-to-one constraints are imposed independently at each spatial location to ensure physical interpretability of abundances. 
{\color{black}In practice, the Abundance Non-negativity Constraint (ANC) is enforced using an element-wise absolute operation, and the Abundance Sum-to-One Constraint (ASC) is satisfied by normalizing the abundances along the endmember dimension as follows:}

{\color{black}
\begin{equation}
\label{eq:asc_anc}
a_{i,j}^{(k)} =
\frac{\left| v_{i,j}^{(k)} \right|}
{\sum_{k=1}^{K} \left| v_{i,j}^{(k)} \right| + \epsilon},
\end{equation}
}

{\color{black}where $\epsilon$ is a small constant used to ensure numerical stability. 

The formulation given in Eq.~\ref{eq:asc_anc} maintains stable gradients while producing normalized abundance vectors that preserve relative material proportions, which are crucial for constructing the affinity matrix used to guide the attention mechanism.}

A lightweight decoder reconstructs the input spectra using a generalized mixing model that combines linear and nonlinear interactions, supervised by the spectral angle distance (SAD) loss. Since this unmixing formulation and training objective are identical to \cite{DCNET}, we omit further architectural and mathematical details and focus on how the learned abundances are leveraged for classification.

\subsection{Abundance-Guided Spectral Transformer}
\label{subsec:abu}
 The second branch of our proposed framework uses a spectral transformer encoder for patch-level classification. Each pixel spectrum within $\mathbf{P}_{i,j}$ is treated as a token, resulting in a sequence of $N=p^2$ spectral tokens. The tokens are then projected by a linear embedding layer into a $d$-dimensional latent space and a learnable classification token is prepended to the sequence.

To inject physically meaningful material information into the spectral transformer, we construct an abundance affinity matrix from the unmixing branch. The abundance tensor $\mathbf{v}_{i,j}$ is flattened to obtain per-pixel abundance vectors $\mathbf{v}' \in \mathbb{R}^{N \times K}$, from which pixel-pixel material similarity is computed as:
\begin{equation}
\label{eq:abundance_affinity}
\mathbf{A_F} = \mathbf{v}' \mathbf{v}'^{\top} \in \mathbb{R}^{N \times N}.
\end{equation}
The abundance affinity matrix captures pixel-pixel similarity in material composition since the abundance vector represents the fractional contribution of latent endmembers within a pixel. High affinity values indicate pixels composed of similar materials, while low values correspond to compositionally distinct pixels. This mechanism provides a similarity measure that is robust to spectral redundancy and mixed-pixel effects, making it well suited for guiding contextual attention in remote sensing scenes.

\textcolor{black}{The affinity matrix described in Eq.~\ref{eq:abundance_affinity} is inherently low-rank, with its rank upper-bounded by the number of endmembers $K$. In our formulation, the number of endmembers is equal to the number of semantic classes, resulting in a compact and structured representation of pixel-wise relationships. While a low-rank structure may appear to limit the expressive capacity of the matrix, it instead introduces a physically meaningful inductive bias. Specifically, it constrains the attention mechanism to model interactions that are consistent with sub-pixel material compositions, thereby suppressing spurious correlations caused by spectral redundancy and noise. Moreover, since $A_F$ is incorporated as an additive bias term in the attention computation (Eq.~\ref{eq:affinity_attention}), the model retains the full flexibility of learned self-attention while being guided by material-aware priors. This design achieves a balance between expressiveness and interpretability, which is particularly beneficial in hyperspectral classification under limited labeled data.}

After padding to match the token sequence length, $\mathbf{A_F}$ is used to guide the transformer attention mechanism.
Specifically, we introduce \emph{abundance-guided multi-head self-attention} \textcolor{black}{\emph{(AGMHSA)}, which integrates abundance information into the attention computation}, where the affinity matrix is added as a bias term to the standard attention logits:
\begin{equation}
\label{eq:affinity_attention}
\mathbf{S}' = \frac{1}{\sqrt{d_h}} \mathbf{Q}\mathbf{K}^{\top} + \lambda \mathbf{A_F},
\end{equation}
where $\mathbf{Q}$ and $\mathbf{K}$ are the projected queries and keys, $d_h$ is the head dimension, and $\lambda$ is a learnable scalar controlling the influence of abundance guidance. This formulation encourages the transformer to attend to pixels with similar sub-pixel material composition, rather than relying solely on spectral similarity. \textcolor{black}{This can be interpreted as a bias-based attention mechanism, where an external similarity prior modulates the attention logits. The affinity matrix is constructed using the inner product of abundance vectors, which reflects the degree of shared material composition between pixels. The additive integration preserves the original query-key similarity while incorporating material-aware relationships, enabling a flexible balance between learned spectral interactions and composition-based guidance through the learnable parameter $\lambda$.}

\subsection{Feature Fusion and Classification}
\label{subsec:feature}
After $L$ transformer layers, the classification token embedding is extracted and normalized using layer normalization to form a global contextual representation. In parallel, the abundance tensor from the unmixing branch is processed by a lightweight convolutional module to obtain a compact global abundance descriptor $\mathbf{g}_{\text{abu}}$. Let $\mathbf{h}_{\text{cls}}$ denote the normalized classification token embedding from the final transformer layer.
The two representations are concatenated and passed through a linear head to produce class logits:
\begin{equation}
\label{eq:feature}
\mathbf{o}_{i,j} = \mathbf{W}_{\text{head}} [\mathbf{h}_{\text{cls}} ; \mathbf{g}_{\text{abu}}] + \mathbf{b}_{\text{head}}.
\end{equation}

\textcolor{black}{The rationale behind using this feature fusion strategy is to combine complementary information from the two branches. The transformer representation $h_{\text{cls}}$ captures global spectral-spatial contextual dependencies, while the abundance descriptor $g_{\text{abu}}$ encodes physically interpretable material composition. To preserve the distinct characteristics of both representations, we concatenate the two, allowing the classifier to jointly leverage contextual features and sub-pixel material information without enforcing premature information mixing. This design enables flexible integration of high-level contextual reasoning with \rev{abundance-derived} cues, leading to more discriminative representations for hyperspectral classification.}

\noindent\textbf{Objective Function:}
AGSA-Net is trained end-to-end using a joint objective that balances physically meaningful unmixing and discriminative classification. Following \cite{DCNET}, the unmixing branch is supervised using the spectral angle distance (SAD) loss, while the transformer classifier is trained using cross-entropy loss. The final training objective is:
\begin{equation}
\mathcal{L} = \mathcal{L}_{\text{CE}} + \alpha \mathcal{L}_{\text{SAD}},
\end{equation}
where $\mathcal{L}_{\text{CE}}$ denotes the cross-entropy classification loss and $\mathcal{L}_{\text{SAD}}$ represents the spectral angle distance loss used to supervise abundance reconstruction. Following DSNet \cite{DCNET}, the weighting coefficient $\alpha$ is fixed to 1 in all experiments.

Algorithm \ref{alg:agsanet} summarizes the AGSA-Net training process, highlighting abundance estimation by the encoder, patch reconstruction by the decoder, and abundance-guided transformer classification.

\begin{algorithm}[t]
\caption{Training of AGSA-Net}
\label{alg:agsanet}
\KwIn{HSI $\mathbf{X}$, labels $y$, patch size $p$, \#endmembers $K$, transformer depth $L$, learning rate $\eta$, loss weight $\alpha$}
\KwOut{Optimized parameters $(\theta_E, \theta_D, \theta_T)$}

\While{not converged}{
  Sample training patch $\mathbf{P}_{i,j}$ with label $y_{i,j}$\;

  {\color{ieeegray}\textit{// Unmixing branch (DSNet-based)}} \\
  $\mathbf{v}_{i,j} \leftarrow f_{\theta_E}(\mathbf{P}_{i,j})$; enforce ANC and ASC\;
  $\mathbf{X}' \leftarrow f_{\theta_D}(\mathbf{v}_{i,j})$\;

  {\color{ieeegray}\textit{// Abundance affinity construction}} \\
  Flatten abundances $\mathbf{v}' \in \mathbb{R}^{N\times K}$\;
  $\mathbf{A_F} \leftarrow \text{pad}(\mathbf{v}'\mathbf{v}'^{\top})$ \textbf{(Eq.~\eqref{eq:abundance_affinity})}\;

  {\color{ieeegray}\textit{// Abundance-guided transformer}} \\
  Initialize tokens $\mathbf{H}^{(0)} = [\mathbf{e}_{cls}; \mathbf{E}] + \mathbf{PE}$\;
  \For{$\ell = 0$ to $L-1$}{
    $\mathbf{H}^{(\ell+1)} \leftarrow \text{AGMHSA}(\mathbf{H}^{(\ell)}, \mathbf{A_F})$ \textbf{(Eq.~\eqref{eq:affinity_attention})}\;
  }

  {\color{ieeegray}\textit{// Fusion and prediction}} \\
  $\mathbf{f} \leftarrow [\,LN(\mathbf{H}^{(L)}_{cls});\, f_{abu}(\mathbf{v}_{i,j})\,]$\;
  $\mathbf{o}_{i,j} \leftarrow \mathbf{W}_{head}\mathbf{f} + \mathbf{b}_{head}$\;

  {\color{ieeegray}\textit{// Joint optimization}} \\
    $\mathcal{L} \leftarrow \mathcal{L}_{\text{CE}} + \alpha \mathcal{L}_{\text{SAD}}$\;

  Update $(\theta_E, \theta_D, \theta_T)$ using $\nabla \mathcal{L}$ and $\eta$\;
}
\end{algorithm}

\section{Experiments and Results}
\label{sec:experiments}
This section presents the experimental framework, datasets, and performance comparison of the proposed model with existing approaches. 
\subsection{Dataset Details and Experimental Setup}
\label{subsec:exp_setup}
The proposed model was evaluated on three benchmark hyperspectral image datasets: Indian Pines, Augsburg, and Berlin. These datasets offer diverse spectral and spatial properties for robust performance assessment. Their key characteristics are summarized in Table ~\ref{tab:datasets}.

\subsubsection*{Indian Pines}
The Indian Pines dataset was collected by the Airborne Visible/Infrared Imaging Spectrometer (AVIRIS) sensor over Northwestern Indiana, USA. It consists of $145 \times 145$ pixels with 224 spectral bands in the wavelength range of 0.4-2.5 $\mu$m. After removing water absorption bands, 200 bands remain. The scene contains 16 different land-cover classes, including agricultural fields, forests, and grasslands.
\subsubsection*{Augsburg}
The Augsburg dataset was obtained using the HySpex sensor along with TerraSAR-X Synthetic Aperture Radar (SAR) imagery and a lidar-derived Digital Surface Model (DSM) over the city of Augsburg in Germany at 30 m spatial resolution. It consists of $332\times 485$ pixels with 180 spectral bands ranging from 400 to 2500 nm and comprises 7 urban materials classes such as vegetation, water, roads, bare soil, and buildings. The SAR data provide detailed backscatter information, and the DSM captures surface variation of heights. 

\subsubsection*{Berlin}
The Berlin dataset was collected over the city of Berlin using the EnMap hyperspectral satellite sensor, along with TerraSAR-X SAR data and a lidar-generated DSM. It consists of $1723 \times 476$ pixels with 244 spectral bands and 30 m spatial resolution covering the wavelength range from 0.4-2.5 $\mu$m and contains 8 land cover classes, such as vegetation, tough surfaces, soil, and multiple building types. The SAR and DSM modalities add texture and elevation cues that improve understanding of complex and urban layouts.

\begin{table}[!t]
\centering
\begin{threeparttable}
\caption{Summary of the HSI datasets used in experiments.}
\label{tab:datasets}
\begin{tabular}{lccc}
\hline
\textbf{Dataset} & \textbf{Indian Pines} & \textbf{Augsburg} & \textbf{Berlin} \\
\hline
Sensor                & AVIRIS  & HySpex & EnMap \\
Wavelength (nm)       & 400--2500 & 400--2500 & 400--2500 \\
Spatial resolution    & 20 m & 30 m & 30 m \\
Spatial size          & 145$\times$145 & 332$\times$485 & 1723$\times$476 \\
Spectral bands        & 200 & 180 & 244 \\
Classes               & 16 & 7 & 8 \\
Additional modalities & None & SAR + DSM & SAR + DSM \\
Data type             & Aerial & Aerial & Aerial \\
\hline
\end{tabular}
\vspace{0.2cm} 
\begin{tablenotes}
\footnotesize
\item \textcolor{black}{SAR: Synthetic Aperture Radar, DSM: Digital Surface Model.}
\end{tablenotes}
\end{threeparttable}
\end{table}
Although the Augsburg and Berlin datasets include additional modalities such as Synthetic Aperture Radar (SAR) and Digital Surface Model (DSM), this study uses only the hyperspectral images (HSI) for training and evaluation. The SAR and DSM data are listed in Table~\ref{tab:datasets} to describe the complete dataset characteristics but are not used in the experiments.

\noindent\textbf{Implementation Details:} Training is performed using the \textit{AdamW} optimizer with learning rate $\eta = 1\times10^{-3}$ and weight decay $1\times10^{-4}$. 
A batch size of 64 ensures stable and efficient optimization. Patch sizes are set to $7\times7$ for the Indian Pines and Augsburg datasets, and $5\times5$ for the Berlin dataset following \cite{DCNET}. Model training and evaluation are conducted on a workstation equipped with an NVIDIA GeForce RTX~3090 GPU with 24~GB memory. 
To ensure a fair comparison with existing work, we strictly followed the experimental protocols used in \cite{DCNET}. We used the same training, validation, and test partitions in all experiments, and no additional resampling strategy was used to ensure that the reported results are evaluated under identical data partitions. All models are trained for 500 epochs, consistent with the training strategy in \cite{DCNET}. \rev{ To quantitatively evaluate the proposed AGSA-Net, we compare it with representative CNN-, transformer-, graph-, and unmixing-based methods, including 2-D CNN \cite{DL}, 3-D CNN \cite{3DCNN}, GRU \cite{gru}, ViT \cite{ViT}, MorphConv \cite{MorphCONVO}, SSFTT \cite{SSFTT}, WFCG \cite{WFCG}, DSNet \cite{DCNET}, CACFTNet \cite{CACFTNet}, GraphGST \cite{GraphGST}, MambaHSI+ \cite{MambaHSI+}, and S2Mamba \cite{S2Mamba}, results on all datasets were directly reported from DSNet \cite{DCNET}. All methods are evaluated under identical experimental settings using Overall Accuracy (OA), Average Accuracy (AA), and the Kappa coefficient. For CACFTNet~\cite{CACFTNet}, GraphGST~\cite{GraphGST}, and S2Mamba~\cite{S2Mamba}, the Indian Pines results were directly reported from their
respective original papers, while the Augsburg and Berlin
results were obtained using the official implementations under
the same data partitions provided by DSNet \cite{DCNET}, since these methods were not originally evaluated on those datasets. For MambaHSI+ \cite{MambaHSI+}, all experiments were conducted using the official implementation with the same data splits from DSNet \cite{DCNET}.} During training, model checkpoints are saved based on classification performance, and the best-performing model is selected for the final evaluation on the test set. To eliminate randomness introduced by initialization and data shuffling, a fixed random seed is used in all experiments.

\subsection{Quantitative Performance Comparison}
\label{subsec:qpc}
\rev{
Tables~\ref{tab:augsburg_classwise}, \ref{tab:berlin_classwise}, and \ref{tab:indian_pines_classwise} present the quantitative comparison of AGSA-Net with existing methods. AGSA-Net achieves competitive performance across all datasets and provides better overall results on the heterogeneous Augsburg and Berlin scenes. On Indian Pines, although some recent methods obtain higher OA, AGSA-Net still outperforms DSNet and shows improved class-wise performance on several spectrally similar crop categories.
The class-wise results further show that AGSA-Net benefits classes affected by spectral overlap and mixed pixels. Compared with DSNet, AGSA-Net improves Corn-notill, Soybean-mintill, Woods, Grass-trees, and Soybean-notill on Indian Pines; Industrial Area, Low Plants, Allotment, Residential Area, and Forest on Augsburg; and Low Plants, Soil, Residential Area, and Forest on Berlin. The improvement can be attributed to the role of AGMHSA. Conventional self-attention forms contextual relationships mainly from learned spectral-feature similarity. This is especially helpful in urban and boundary regions, where spectrally similar pixels may belong to different semantic classes, and where raw spectral similarity alone may produce unreliable attention patterns. Therefore, the gains on Industrial Area, Low Plants, Residential Area, Soil, and mixed agricultural classes support the claim that abundance guidance improves contextual reasoning under spectral ambiguity and mixed-pixel conditions.
}

\begin{figure*}[t]
    \centering
    \includegraphics[width=0.47\textwidth]{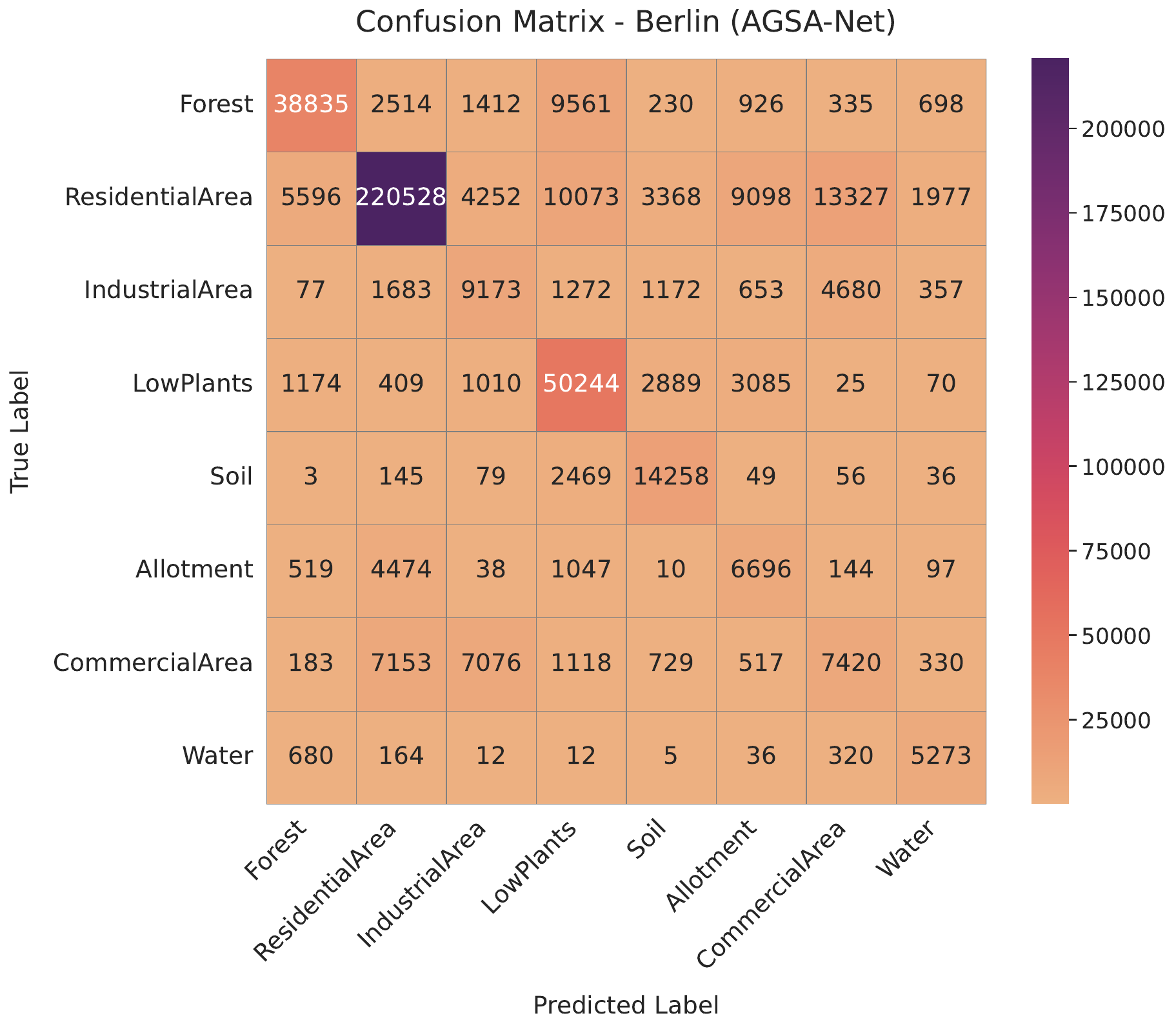}
    \hfill
    \includegraphics[width=0.47\textwidth]{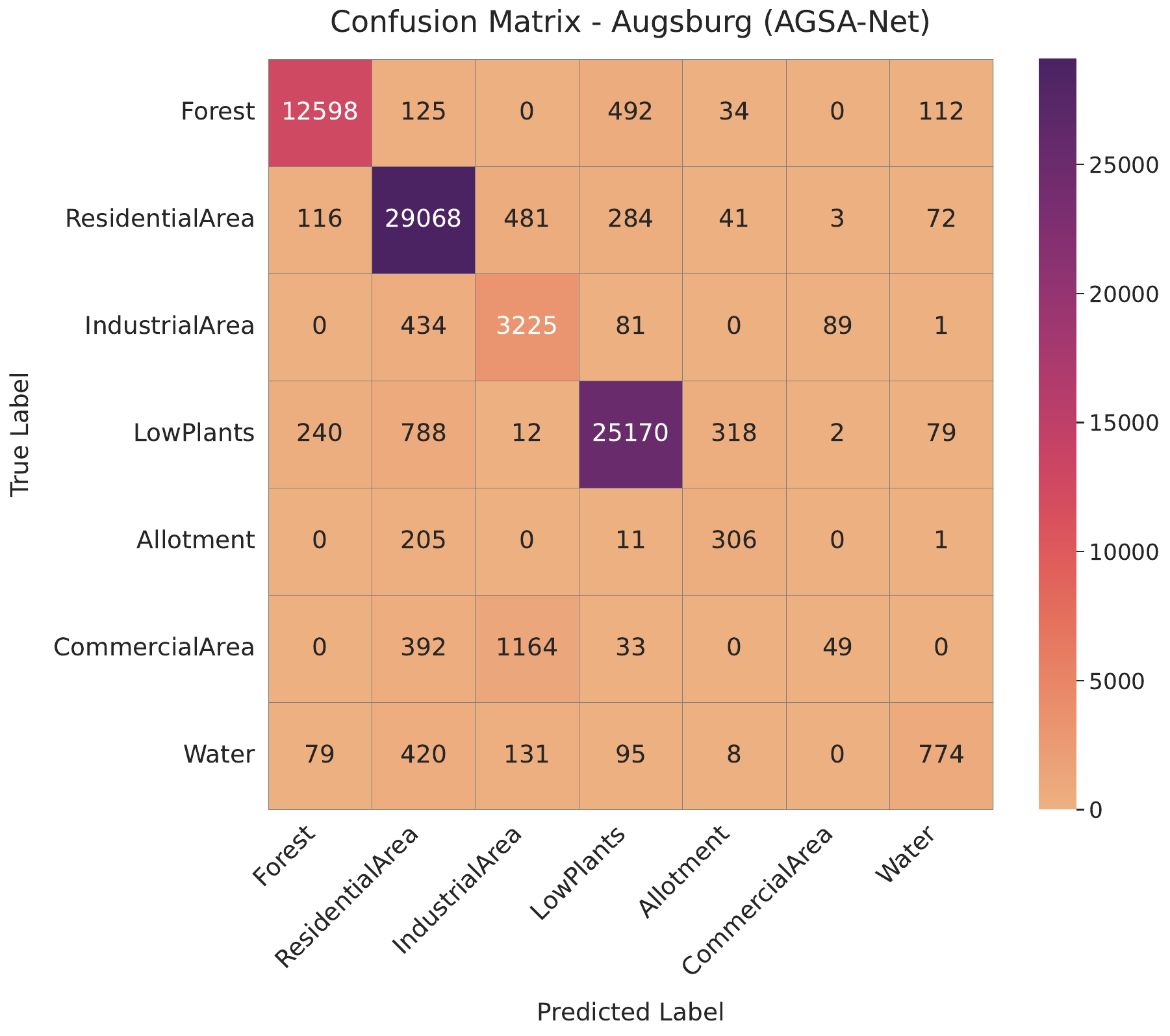}
    \caption{\rev{Confusion matrices obtained using AGSA-Net on the Berlin and Augsburg datasets.}}
    \label{fig:confusion_matrices}
\end{figure*}

\rvn{
However, AGSA-Net does not improve all classes uniformly. Commercial Area and Water remain difficult on Augsburg, while Industrial Area, Allotment, Commercial Area, and Water show lower performance than DSNet on Berlin. These errors are mainly caused by severe class overlap, limited training samples, and shared materials among urban categories such as commercial, industrial, and residential regions, as shown in Fig. \ref{fig:confusion_matrices}. Thus, the proposed attention mechanism improves contextual aggregation but cannot fully resolve extreme class imbalance or highly overlapping semantic classes.
}

\begin{figure*}[t]
\centering
\scriptsize
\begin{minipage}{0.19\textwidth}
\centering
False Color Image\\[2pt]
\includegraphics[width=\linewidth]{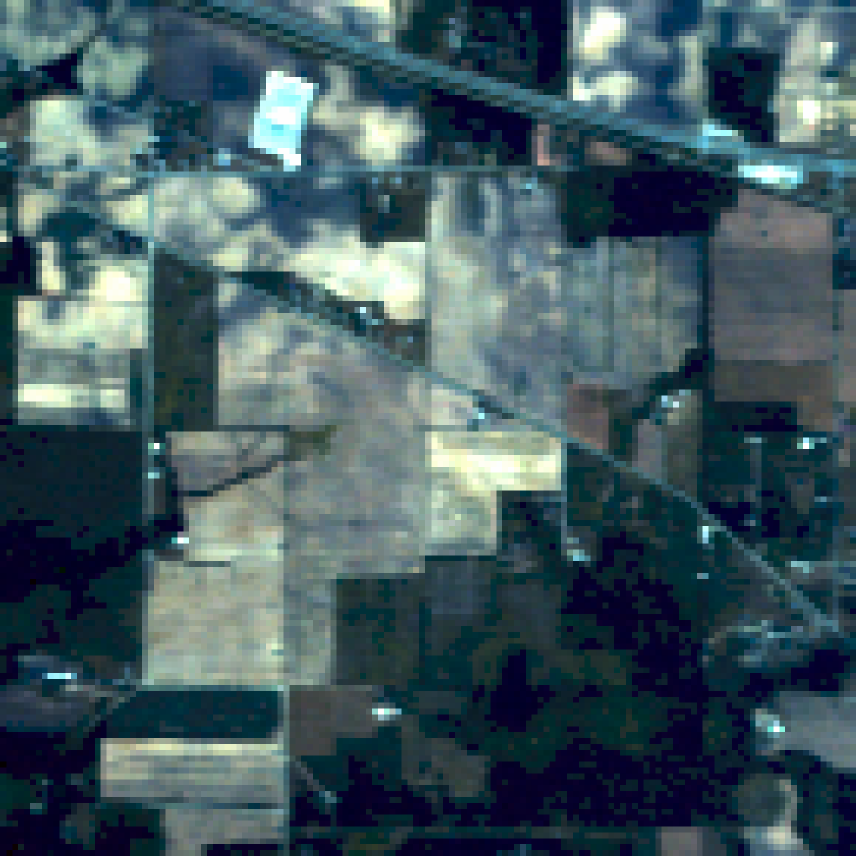}
\end{minipage}
\hfill
\begin{minipage}{0.19\textwidth}
\centering
Ground Truth\\[2pt]
\includegraphics[width=\linewidth]{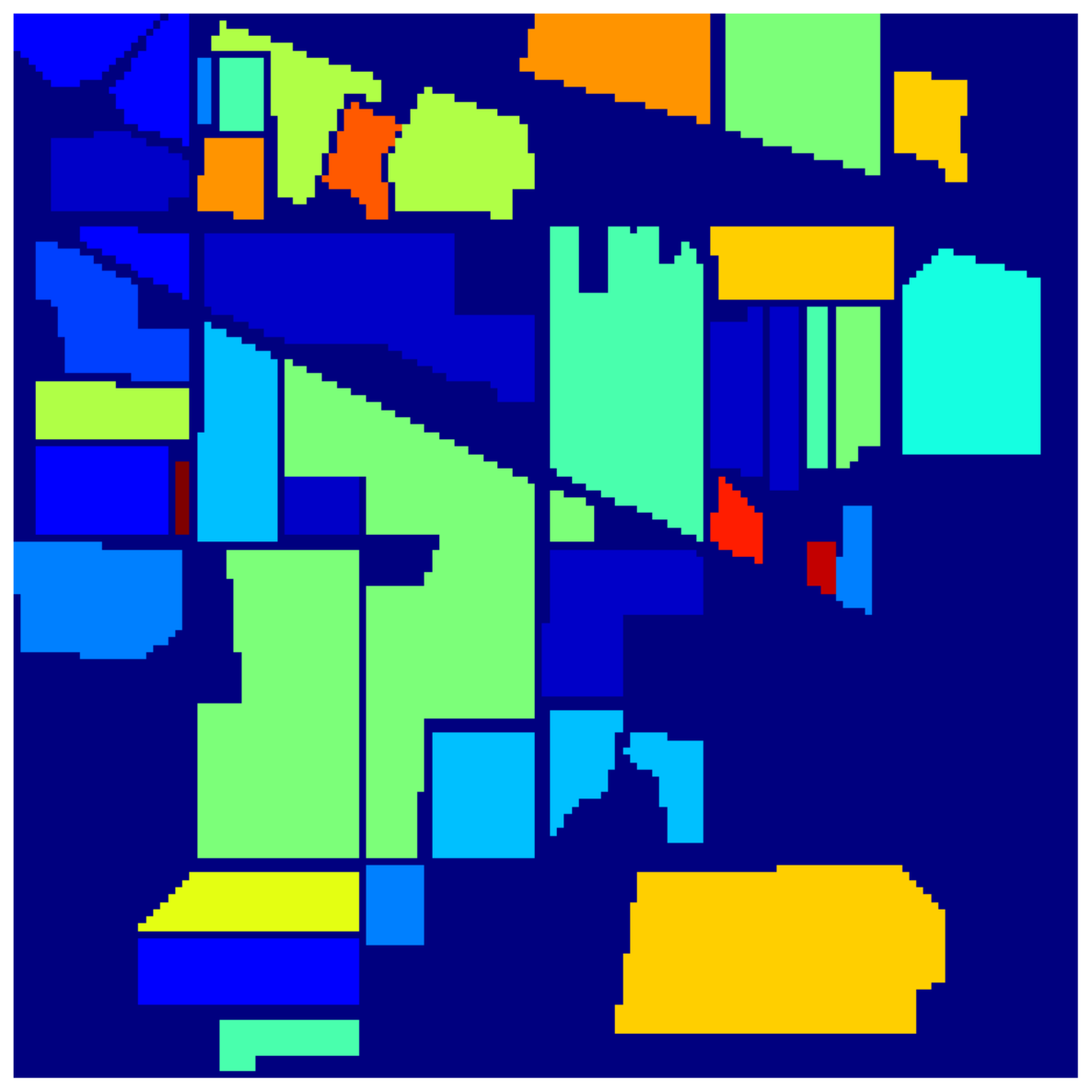}
\end{minipage}
\hfill
\begin{minipage}{0.19\textwidth}
\centering
Transformer\\[2pt]
\includegraphics[width=\linewidth]{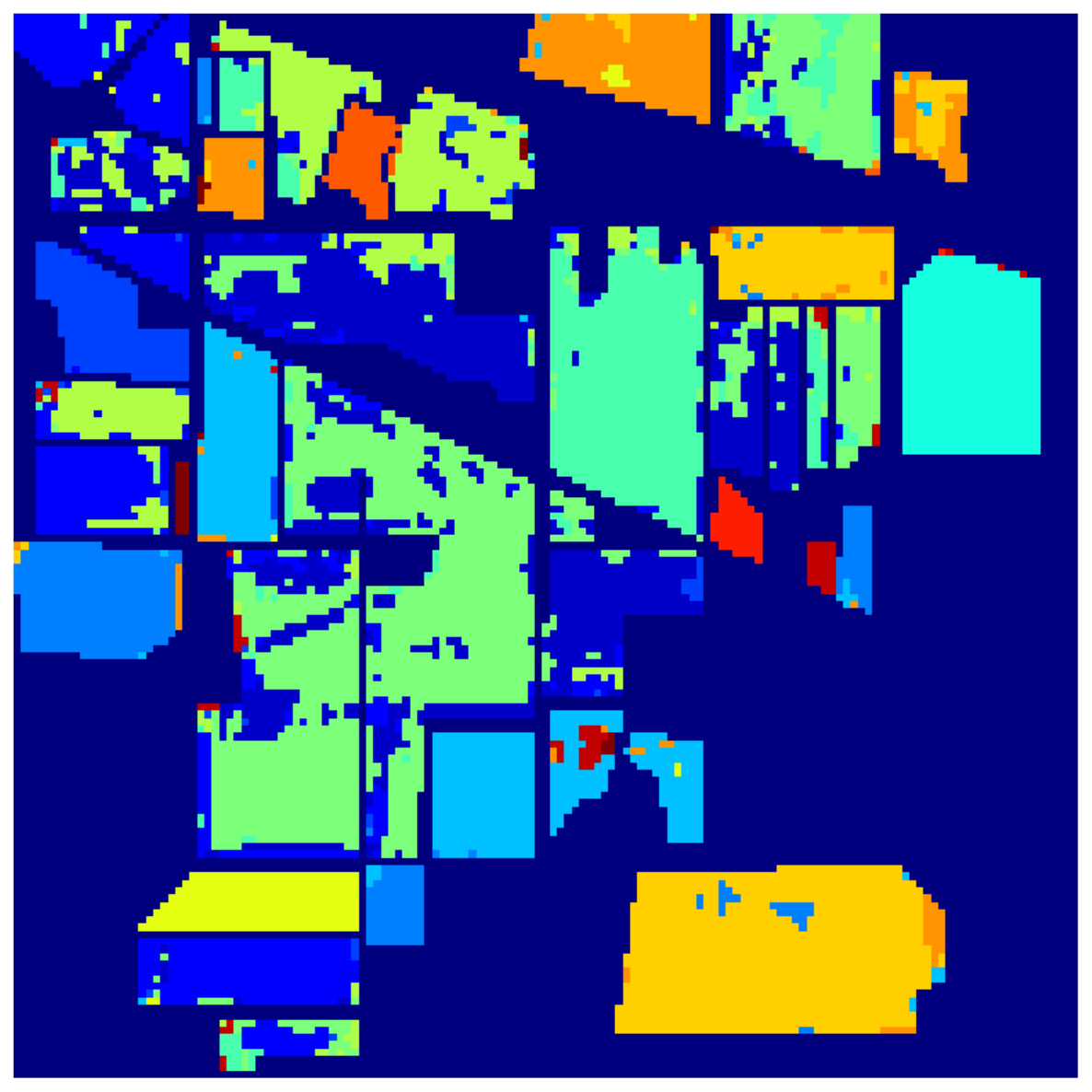}
\end{minipage}
\hfill
\begin{minipage}{0.19\textwidth}
\centering
DSNet~\cite{DCNET}\\[2pt]
\includegraphics[width=\linewidth]{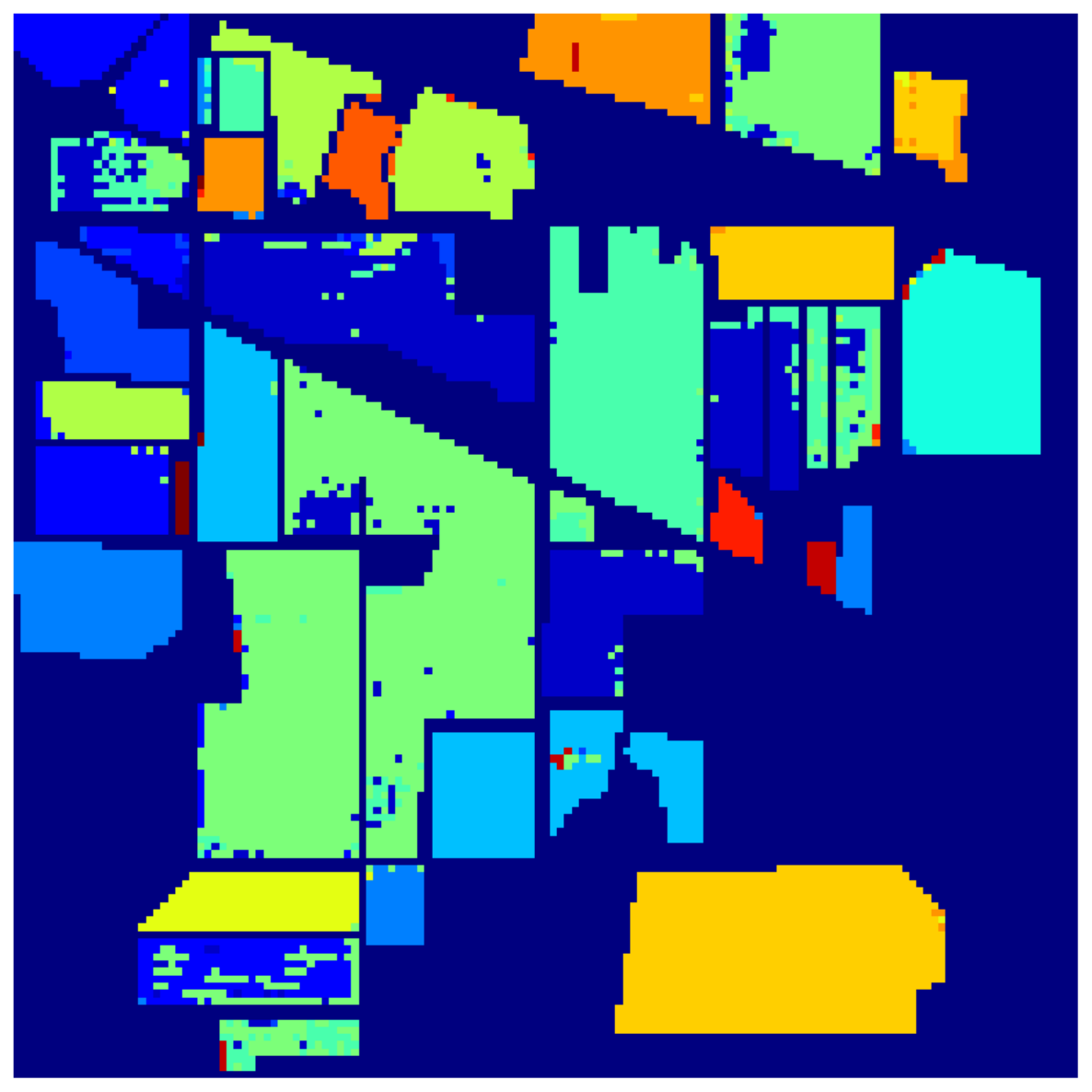}
\end{minipage}
\hfill
\begin{minipage}{0.19\textwidth}
\centering
AGSA-Net (Ours)\\[2pt]
\includegraphics[width=\linewidth]{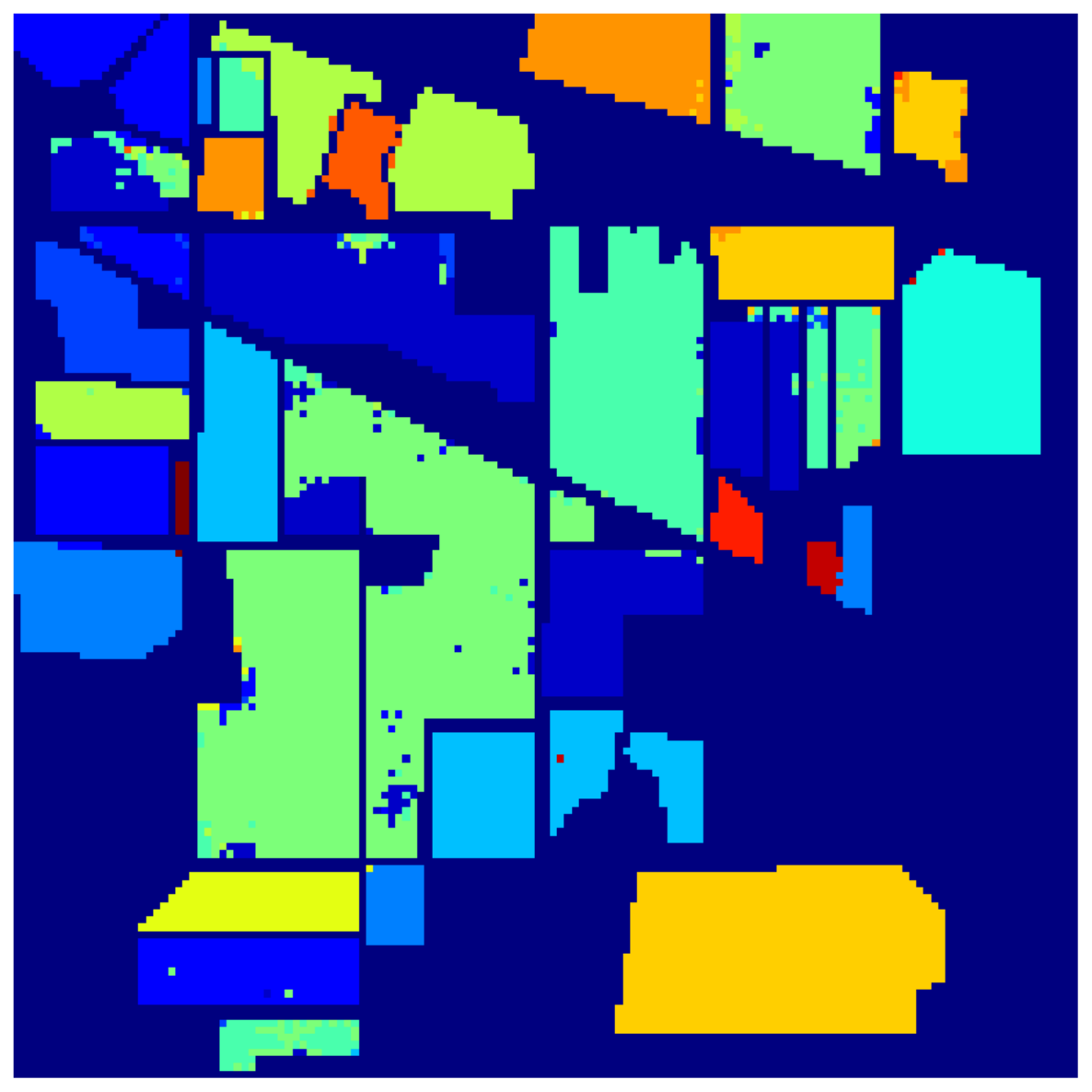}
\end{minipage}
\centerline{\includegraphics[width=\textwidth]{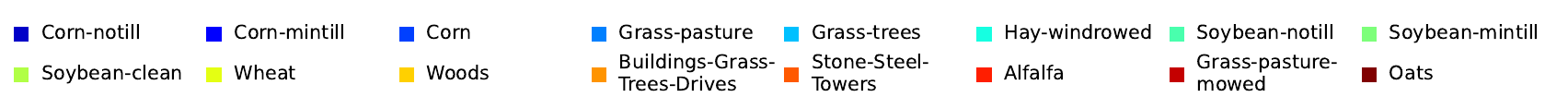}}
\caption{False-color image, ground truth and classification maps obtained using the Transformer, DSNet, and AGSA-Net models on the Indian Pines dataset.}
\label{fig:ip_collage}
\end{figure*}

\begin{figure*}[t]
\centering
\scriptsize
\begin{minipage}{0.19\textwidth}
\centering
False Color Image\\[2pt]
\includegraphics[width=\linewidth]{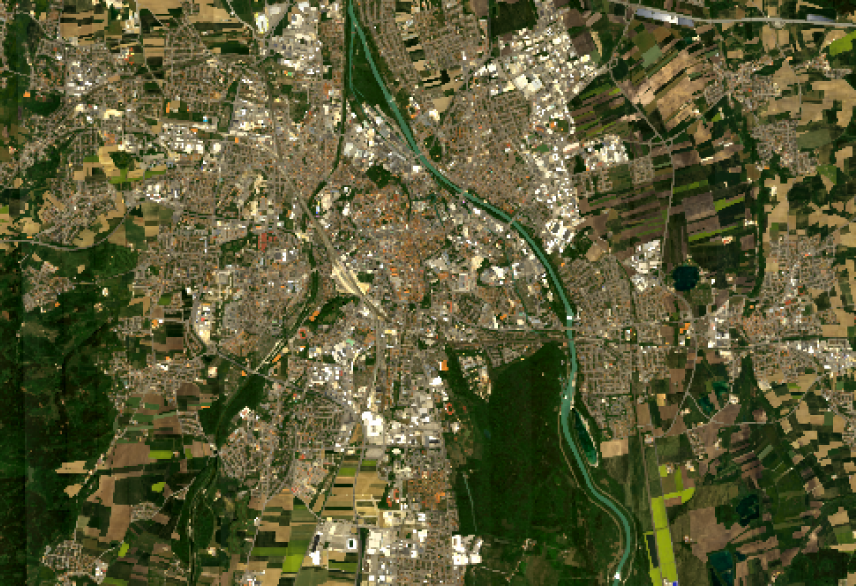}
\end{minipage}
\hfill
\begin{minipage}{0.19\textwidth}
\centering
Ground Truth\\[2pt]
\includegraphics[width=\linewidth]{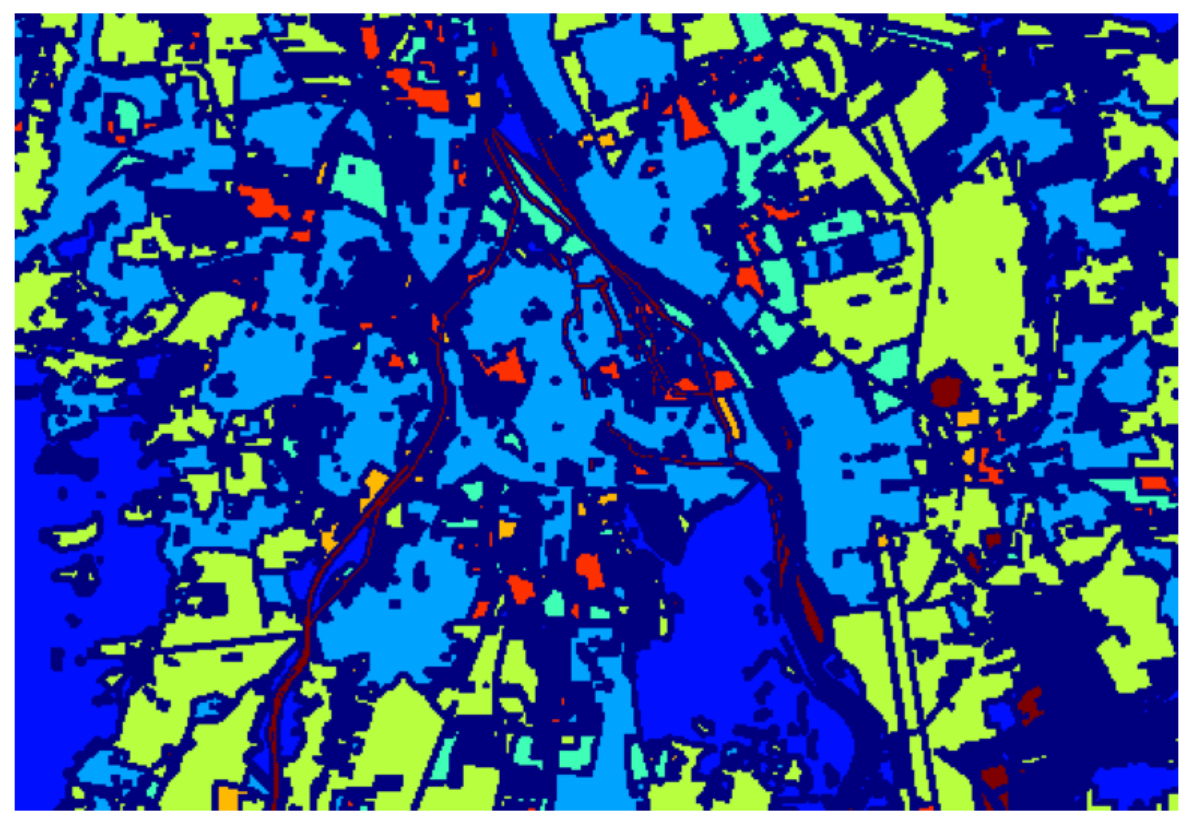}
\end{minipage}
\hfill
\begin{minipage}{0.19\textwidth}
\centering
Transformer\\[2pt]
\includegraphics[width=\linewidth]{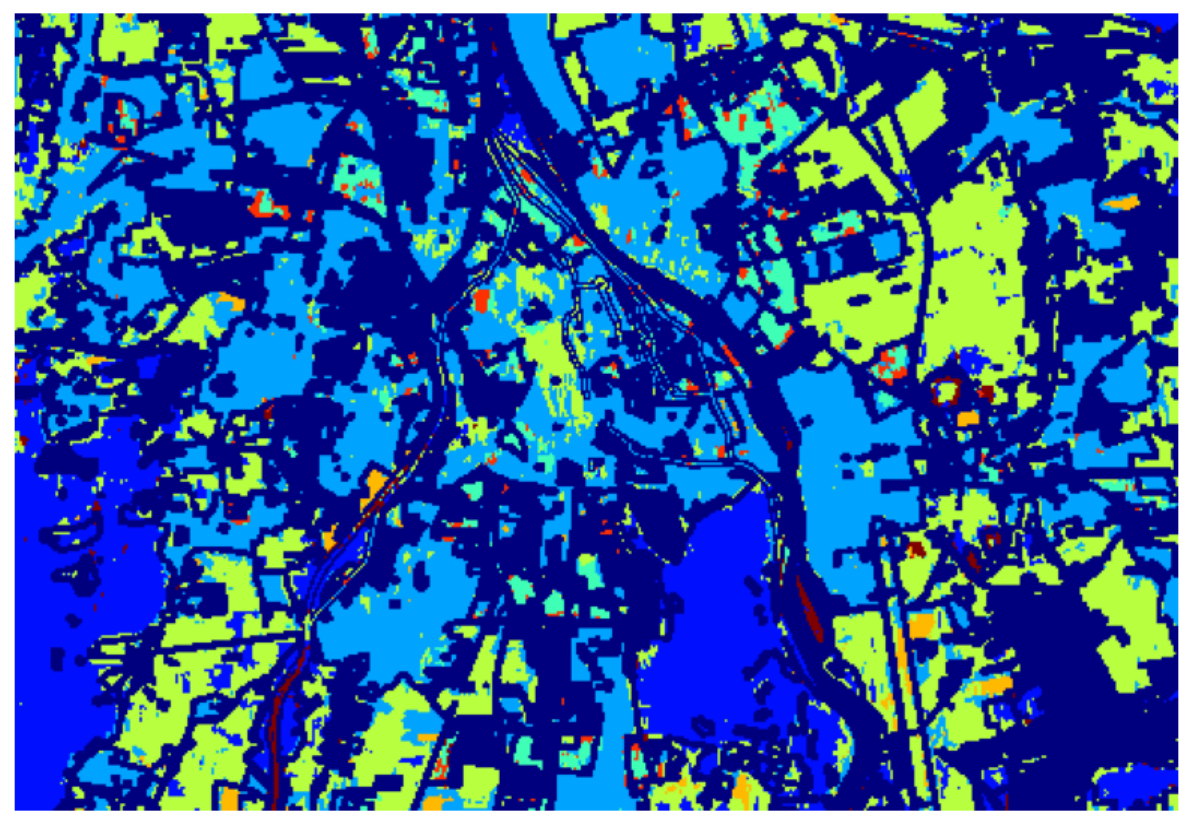}
\end{minipage}
\hfill
\begin{minipage}{0.19\textwidth}
\centering
DSNet~\cite{DCNET}\\[2pt]
\includegraphics[width=\linewidth]{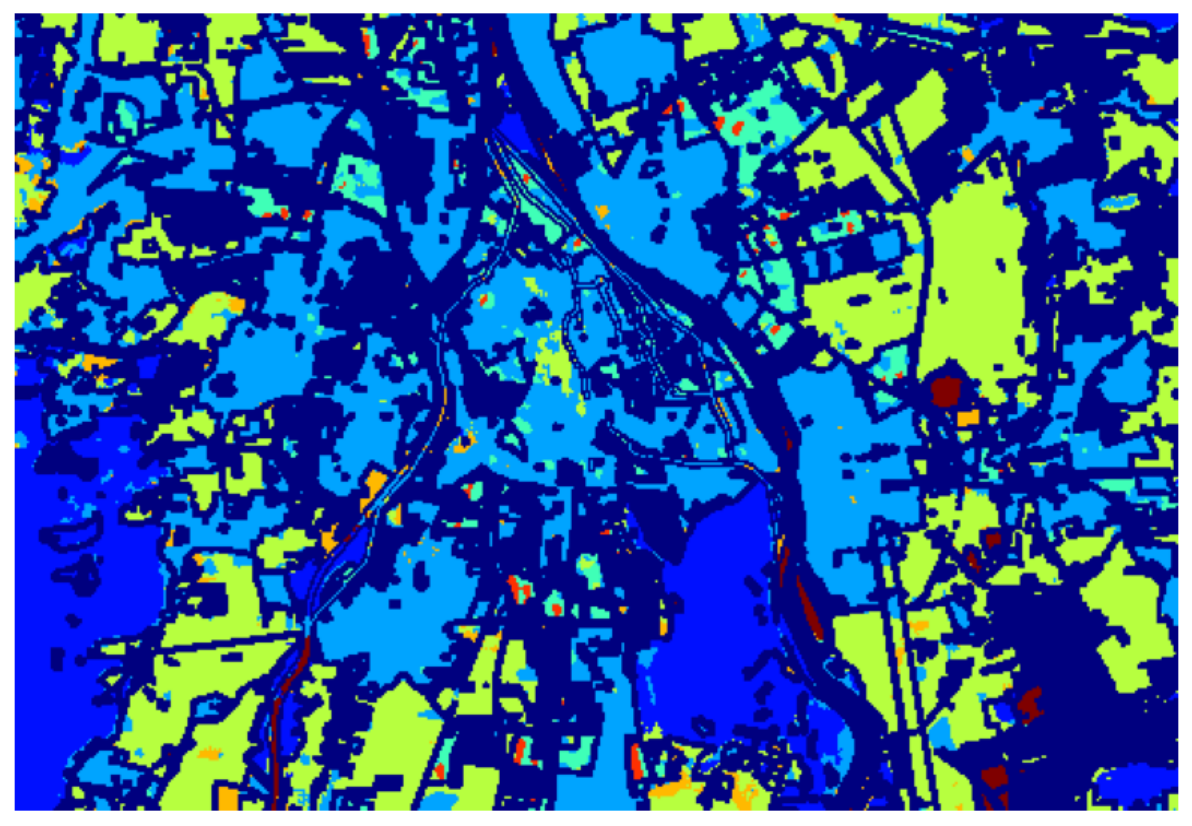}
\end{minipage}
\hfill
\begin{minipage}{0.19\textwidth}
\centering
AGSA-Net (Ours)\\[2pt]
\includegraphics[width=\linewidth]{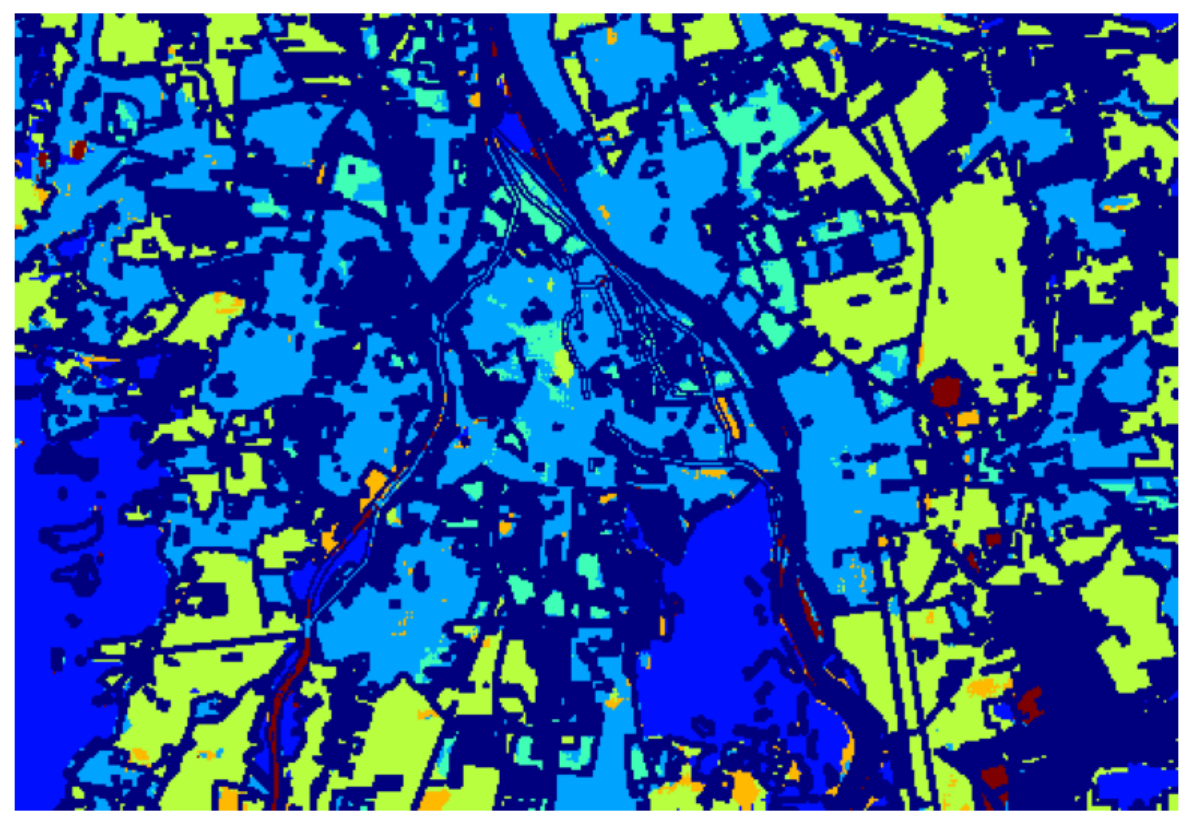}
\end{minipage}
\centerline{\includegraphics[width=\textwidth]{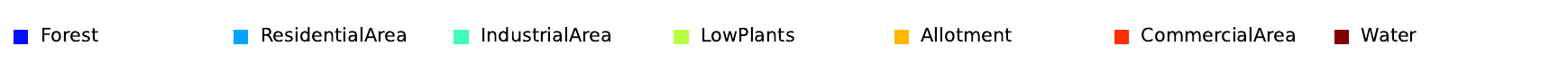}}
\caption{False-color image, ground truth, and classification maps obtained using the Transformer, DSNet, and AGSA-Net models on the Augsburg dataset.}
\label{fig:ag_collage}
\end{figure*}

\begin{figure}[t]
\centering
\berlinsize
\begin{minipage}{0.18\linewidth}
\centering
False Color Image\\[2pt]
\includegraphics[width=\linewidth]{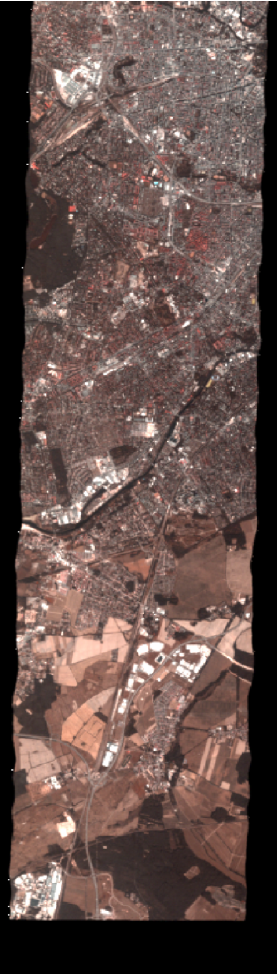}
\end{minipage}
\hfill
\begin{minipage}{0.18\linewidth}
\centering
Ground Truth\\[2pt]
\includegraphics[width=\linewidth]{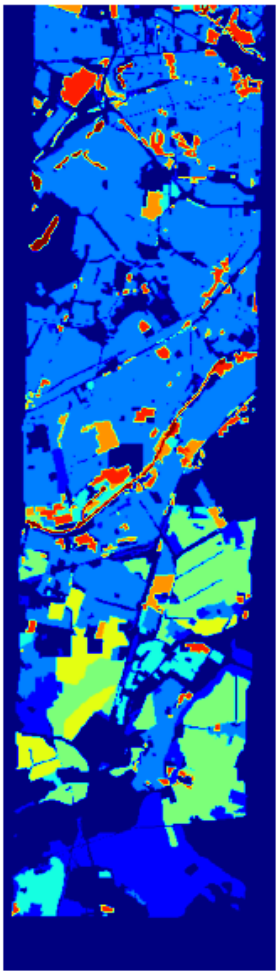}
\end{minipage}
\hfill
\begin{minipage}{0.18\linewidth}
\centering
Transformer\\[2pt]
\includegraphics[width=\linewidth]{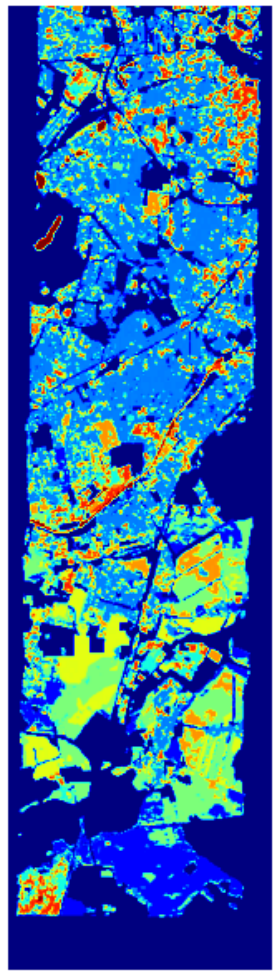}
\end{minipage}
\hfill
\begin{minipage}{0.18\linewidth}
\centering
DSNet~\cite{DCNET}\\[2pt]
\includegraphics[width=\linewidth]{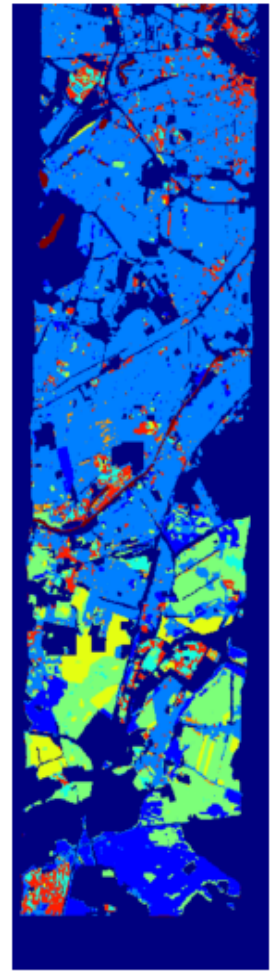}
\end{minipage}
\hfill
\begin{minipage}{0.18\linewidth}
\centering
AGSA-Net (Ours)\\[2pt]
\includegraphics[width=\linewidth]{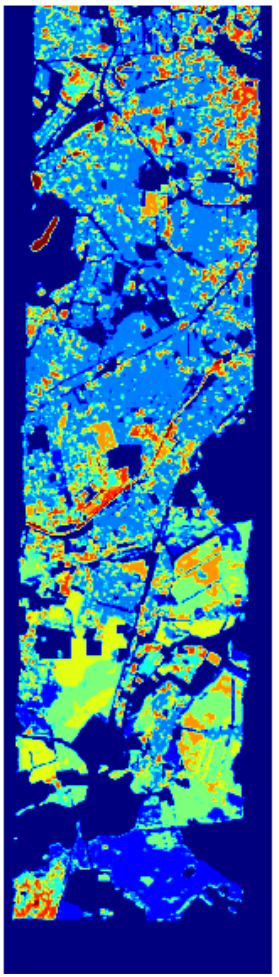}
\end{minipage}
\centerline{\includegraphics[width=\columnwidth]{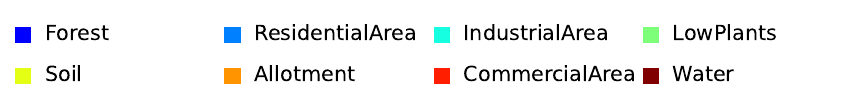}}
\caption{False-color image, ground truth and classification maps obtained using the Transformer, DSNet, and AGSA-Net models on the Berlin dataset.}
\label{fig:br_collage}
\end{figure}

\rvn{
Qualitative results in Figs.~\ref{fig:ip_collage}, \ref{fig:ag_collage}, and \ref{fig:br_collage} further support these observations. In Augsburg and Berlin, AGSA-Net produces more spatially coherent predictions in urban regions and reduces fragmented errors near class boundaries. Nevertheless, residual errors are still observed among spectrally similar urban surface classes in Berlin and in minority classes with limited training samples in Augsburg. These observations indicate that abundance-guided attention improves contextual discrimination and reduces class confusion, but does not completely eliminate ambiguity in highly heterogeneous urban scenes.
}

\rev{The computational complexity of AGSA-Net is summarized in Table~\ref{tab:complexity}. Compared with lightweight models such as DSNet and SSFTT, AGSA-Net introduces a moderate increase in FLOPs and inference time due to the
additional contextual modeling introduced by the transformer branch and abundance-guided attention mechanism. In particular, AGSA-Net exhibits higher inference time than DSNet and SSFTT, indicating that the proposed architecture does not primarily optimize for minimal computational overhead. However, compared with larger Transformer-based models such as CACFTNet and ViT, AGSA-Net requires substantially fewer FLOPs and achieves faster inference while maintaining a compact parameter footprint. This efficiency comes from its compact token representation and guided attention design, which avoids the high cost of large embedding dimensions and full global self-attention. Therefore, AGSA-Net should be viewed as a practical trade-off between improved representation capability and moderate computational cost.}

\begin{table*}[t]
\centering
\caption{Class-wise classification accuracy (\%) on the Augsburg dataset}
\label{tab:augsburg_classwise}
\scriptsize
\setlength{\tabcolsep}{2pt}
\begin{tabular*}{\textwidth}{@{\extracolsep{\fill}}lcc|ccccccccccccc}
\hline
\textbf{Class Name} & \textbf{Train} & \textbf{Test} &
\shortstack[c]{\textbf{2-D CNN}\\\cite{DL}} &
\shortstack[c]{\textbf{3-D CNN}\\\cite{3DCNN}} &
\shortstack[c]{\textbf{GRU}\\\cite{gru}} &
\shortstack[c]{\textbf{ViT}\\\cite{ViT}} &
\shortstack[c]{\textbf{MorphConv}\\\cite{MorphCONVO}} &
\shortstack[c]{\textbf{SSFTT}\\\cite{SSFTT}} &
\shortstack[c]{\textbf{WFCG}\\\cite{WFCG}} &
\shortstack[c]{\textbf{DSNet}\\\cite{DCNET}} &
\shortstack[c]{\textcolor{black}{\textbf{CACFTNet}}\\\textcolor{black}{\cite{CACFTNet}}} &
\shortstack[c]{\textcolor{black}{\textbf{GraphGST}}\\\textcolor{black}{\cite{GraphGST}}} &
\shortstack[c]{\textcolor{black}{\textbf{MambaHSI+}}\\\textcolor{black}{\cite{MambaHSI+}}} &
\shortstack[c]{\textcolor{black}{\textbf{S2Mamba}}\\\textcolor{black}{\cite{S2Mamba}}} &
\shortstack[c]{\textbf{AGSA-Net}\\(Ours)} \\
\hline
Forest & 146 & 13361 & 85.76 & 93.02 & 81.10 & 82.44 & 93.21 & 95.52 & 91.65 & 93.19 &
\textcolor{black}{\textbf{98.23}} & \textcolor{black}{87.89} & \textcolor{black}{95.42} & \textcolor{black}{87.07} & 94.66 \\
Residential Area & 264 & 30065 & 87.44 & 92.68 & 84.51 & 88.77 & 95.83 & 96.39 & \textbf{98.96} & 95.85 &
\textcolor{black}{96.60} & \textcolor{black}{92.89} & \textcolor{black}{96.86} & \textcolor{black}{93.68} & 96.95 \\
Industrial Area & 21 & 3830 & 69.56 & 21.46 & \textbf{84.41} & 79.27 & 32.69 & 25.93 & 50.97 & 62.85 &
\textcolor{black}{66.40} & \textcolor{black}{42.64} & \textcolor{black}{64.88} & \textcolor{black}{63.55} & 73.68 \\
Low Plants & 248 & 26609 & 81.39 & 78.29 & 82.25 & 80.51 & 88.75 & 87.02 & 87.33 & 91.28 &
\textcolor{black}{92.86} & \textcolor{black}{82.89} & \textcolor{black}{88.76} & \textcolor{black}{93.47} & \textbf{93.99} \\
Allotment & 52 & 523 & 41.68 & 58.70 & 32.10 & 33.35 & 63.67 & 45.70 & 52.35 & 66.92 &
\textcolor{black}{68.64} & \textcolor{black}{53.35} & \textcolor{black}{55.64} & \textcolor{black}{43.98} & \textbf{69.41} \\
Commercial Area & 7 & 1638 & 14.10 & 9.46 & 5.56 & 12.39 & \textbf{15.81} & 11.78 & 1.05 & 11.05 &
\textcolor{black}{\textbf{16.12}} & \textcolor{black}{10.32} & \textcolor{black}{10.93} & \textcolor{black}{12.45} & 4.58 \\
Water & 23 & 1507 & 35.83 & 9.36 & 19.18 & 19.04 & 17.25 & 26.28 & 33.92 & \textbf{49.24} &
\textcolor{black}{22.89} & \textcolor{black}{40.28} & \textcolor{black}{36.03} & \textcolor{black}{24.29} & 46.05 \\
\hline
OA (\%) & & & 81.33 & 80.67 & 79.85 & 81.04 & 86.39 & 86.05 & 87.66 & 89.30 &
\textcolor{black}{90.78} & \textcolor{black}{83.08} & \textcolor{black}{88.98} & \textcolor{black}{87.58} & \textbf{91.26} \\
AA (\%) & & & 59.40 & 51.85 & 55.59 & 56.58 & 58.17 & 55.52 & 59.46 & 67.20 &
\textcolor{black}{65.96} & \textcolor{black}{58.61} & \textcolor{black}{64.07} & \textcolor{black}{59.78} & \textbf{68.47} \\
Kappa (\%) & & & 73.27 & 72.01 & 71.41 & 72.96 & 80.25 & 79.59 & 82.07 & 84.62 &
\textcolor{black}{86.77} & \textcolor{black}{75.51} & \textcolor{black}{84.14} & \textcolor{black}{82.29} & \textbf{87.46} \\
\hline
\end{tabular*}
\end{table*}

\begin{table*}[t]
\centering
\caption{Class-wise classification accuracy (\%) on the Berlin dataset}
\label{tab:berlin_classwise}
\scriptsize
\setlength{\tabcolsep}{2pt}
\begin{tabular*}{\textwidth}{@{\extracolsep{\fill}}lcc|ccccccccccccc}
\hline
\textbf{Class} & \textbf{Train} & \textbf{Test} &
\shortstack[c]{\textbf{2-D CNN}\\\cite{DL}} &
\shortstack[c]{\textbf{3-D CNN}\\\cite{3DCNN}} &
\shortstack[c]{\textbf{GRU}\\\cite{gru}} &
\shortstack[c]{\textbf{ViT}\\\cite{ViT}} &
\shortstack[c]{\textbf{MorphConv}\\\cite{MorphCONVO}} &
\shortstack[c]{\textbf{SSFTT}\\\cite{SSFTT}} &
\shortstack[c]{\textbf{WFCG}\\\cite{WFCG}} &
\shortstack[c]{\textbf{DSNet}\\\cite{DCNET}} &
\shortstack[c]{\textcolor{black}{\textbf{CACFTNet}}\\\textcolor{black}{\cite{CACFTNet}}} &
\shortstack[c]{\textcolor{black}{\textbf{GraphGST}}\\\textcolor{black}{\cite{GraphGST}}} &
\shortstack[c]{\textcolor{black}{\textbf{MambaHSI+}}\\\textcolor{black}{\cite{MambaHSI+}}} &
\shortstack[c]{\textcolor{black}{\textbf{S2Mamba}}\\\textcolor{black}{\cite{S2Mamba}}} &
\shortstack[c]{\textbf{AGSA-Net}\\(Ours)} \\
\hline
Forest & 443 & 54511 & 77.16 & 73.10 & 75.55 & 66.74 & 55.24 & 47.84 & 76.46 & 75.44 &
\textcolor{black}{73.55} & \textcolor{black}{84.93} & \textcolor{black}{81.42} & \textcolor{black}{70.24} & \textbf{76.51} \\
Residential & 423 & 268219 & 66.33 & 55.27 & 57.60 & 57.86 & 72.47 & 73.58 & 68.80 & 74.69 &
\textcolor{black}{84.41} & \textcolor{black}{72.97} & \textcolor{black}{81.79} & \textcolor{black}{74.05} & \textbf{79.31} \\
Industrial & 499 & 19067 & 21.37 & 43.92 & \textbf{73.13} & 55.85 & 26.25 & 30.59 & 42.06 & 54.91 &
\textcolor{black}{38.76} & \textcolor{black}{53.75} & \textcolor{black}{32.35} & \textcolor{black}{41.51} & 49.26 \\
LowPlants & 376 & 58906 & 69.11 & 68.76 & 71.70 & 76.25 & 74.27 & 84.73 & 67.58 & 81.79 &
\textcolor{black}{62.63} & \textcolor{black}{86.15} & \textcolor{black}{73.94} & \textcolor{black}{75.77} & \textbf{89.26} \\
Soil & 331 & 17095 & 68.15 & 84.05 & 82.95 & 80.04 & 84.88 & 92.20 & \textbf{97.91} & 74.19 &
\textcolor{black}{69.01} & \textcolor{black}{83.18} & \textcolor{black}{71.36} & \textcolor{black}{76.02} & 80.51 \\
Allotment & 280 & 13025 & 58.52 & 32.48 & 22.30 & 64.17 & 61.21 & 49.21 & \textbf{72.80} & 63.71 &
\textcolor{black}{29.25} & \textcolor{black}{77.47} & \textcolor{black}{33.65} & \textcolor{black}{64.19} & 57.83 \\
Commercial & 298 & 24526 & 47.47 & 30.50 & 10.64 & 29.90 & 49.90 & 39.25 & \textbf{53.21} & 30.56 &
\textcolor{black}{33.53} & \textcolor{black}{32.22} & \textcolor{black}{22.18} & \textcolor{black}{34.21} & 24.52 \\
Water & 170 & 6502 & 65.70 & 58.27 & 67.24 & \textbf{78.05} & 65.47 & 63.63 & 71.16 & 67.44 &
\textcolor{black}{65.89} & \textcolor{black}{69.07} & \textcolor{black}{66.46} & \textcolor{black}{74.85} & 63.09 \\
\hline
OA (\%) & & & 64.94 & 57.77 & 59.75 & 60.97 & 67.60 & 68.23 & 68.83 & 72.09 &
\textcolor{black}{73.38} & \textcolor{black}{73.56} & \textcolor{black}{73.58} & \textcolor{black}{70.17} & \textbf{75.31} \\
AA (\%) & & & 59.23 & 55.80 & 57.64 & 63.61 & 61.21 & 60.13 & 68.75 & 65.34 &
\textcolor{black}{57.13} & \textcolor{black}{69.97} & \textcolor{black}{57.89} & \textcolor{black}{63.86} & \textbf{65.04} \\
Kappa (\%) & & & 50.06 & 42.84 & 45.02 & 47.28 & 53.38 & 53.73 & 56.48 & 59.27 &
\textcolor{black}{58.40} & \textcolor{black}{62.05} & \textcolor{black}{59.18} & \textcolor{black}{56.65} & \textbf{63.04} \\
\hline
\end{tabular*}
\end{table*}

\begin{table*}[t]
\centering
\caption{Class-wise classification accuracy (\%) on the Indian Pines dataset}
\label{tab:indian_pines_classwise}
\scriptsize
\setlength{\tabcolsep}{2pt}
\begin{tabular*}{\textwidth}{@{\extracolsep{\fill}}lcc|ccccccccccccc}
\hline
\textbf{Class Name} & \textbf{Train} & \textbf{Test} &
\shortstack[c]{\textbf{2-D CNN}\\\cite{DL}} &
\shortstack[c]{\textbf{3-D CNN}\\\cite{3DCNN}} &
\shortstack[c]{\textbf{GRU}\\\cite{gru}} &
\shortstack[c]{\textbf{ViT}\\\cite{ViT}} &
\shortstack[c]{\textbf{MorphConv}\\\cite{MorphCONVO}} &
\shortstack[c]{\textbf{SSFTT}\\\cite{SSFTT}} &
\shortstack[c]{\textbf{WFCG}\\\cite{WFCG}} &
\shortstack[c]{\textbf{DSNet}\\\cite{DCNET}} &
\shortstack[c]{\textcolor{black}{\textbf{CACFTNet}}\\\textcolor{black}{\cite{CACFTNet}}} &
\shortstack[c]{\textcolor{black}{\textbf{GraphGST}}\\\textcolor{black}{\cite{GraphGST}}} &
\shortstack[c]{\textcolor{black}{\textbf{MambaHSI+}}\\\textcolor{black}{\cite{MambaHSI+}}} &
\shortstack[c]{\textcolor{black}{\textbf{S2Mamba}}\\\textcolor{black}{\cite{S2Mamba}}} &
\shortstack[c]{\textcolor{black}{\textbf{AGSA-Net}}\\\textcolor{black}{(Ours)}} \\
\hline
Corn Notill & 50 & 1384 & 77.24 & 65.25 & 72.33 & 66.55 & 71.53 & 85.62 & \textbf{94.87} & 83.96 &
\textcolor{black}{\textbf{95.88}} & \textcolor{black}{95.81} & \textcolor{black}{74.57} & \textcolor{black}{94.44} & \textcolor{black}{91.04} \\
Corn Mintill & 50 & 784 & 84.57 & 59.57 & 83.16 & 88.52 & 84.57 & 90.43 & 90.69 & 94.90 &
\textcolor{black}{99.74} & \textcolor{black}{98.85} & \textcolor{black}{87.63} & \textcolor{black}{\textbf{100.00}} & \textcolor{black}{96.56} \\
Corn & 50 & 184 & 92.39 & 94.57 & 73.37 & 88.59 & 97.83 & 97.28 & \textbf{100.00} & 99.46 &
\textcolor{black}{\textbf{100.00}} & \textcolor{black}{\textbf{100.00}} & \textcolor{black}{95.65} & \textcolor{black}{\textbf{100.00}} & \textcolor{black}{99.46} \\
Grass Pasture & 50 & 447 & 94.63 & 89.71 & 87.02 & 97.09 & 93.74 & 90.60 & 94.41 & 98.21 &
\textcolor{black}{97.76} & \textcolor{black}{97.09} & \textcolor{black}{87.92} & \textcolor{black}{\textbf{98.43}} & \textcolor{black}{97.32} \\
Grass Trees & 50 & 697 & 78.62 & 95.12 & 81.35 & 85.65 & 92.25 & 95.95 & 99.43 & 97.27 &
\textcolor{black}{\textbf{100.00}} & \textcolor{black}{98.57} & \textcolor{black}{90.24} & \textcolor{black}{\textbf{100.00}} & \textcolor{black}{\textbf{100.00}} \\
Hay Windrowed & 50 & 439 & 92.26 & 98.63 & 94.76 & 98.86 & 98.86 & 98.63 & \textbf{100.00} & 99.54 &
\textcolor{black}{\textbf{100.00}} & \textcolor{black}{99.54} & \textcolor{black}{95.67} & \textcolor{black}{\textbf{100.00}} & \textcolor{black}{98.41} \\
Soybean Notill & 50 & 918 & 79.63 & 79.74 & 84.86 & 93.03 & 73.31 & 85.51 & 82.79 & 95.10 &
\textcolor{black}{\textbf{98.80}} & \textcolor{black}{97.93} & \textcolor{black}{79.96} & \textcolor{black}{98.47} & \textcolor{black}{97.17} \\
Soybean Mintill & 50 & 2418 & 67.74 & 67.70 & 66.50 & 73.95 & 56.95 & 75.27 & 87.68 & 90.07 &
\textcolor{black}{97.89} & \textcolor{black}{94.50} & \textcolor{black}{71.22} & \textcolor{black}{\textbf{98.10}} & \textcolor{black}{94.25} \\
Soybean Clean & 50 & 564 & 77.62 & 72.87 & 64.00 & 75.18 & 81.91 & 78.90 & 86.70 & 94.68 &
\textcolor{black}{\textbf{96.98}} & \textcolor{black}{95.04} & \textcolor{black}{72.52} & \textcolor{black}{95.04} & \textcolor{black}{94.50} \\
Wheat & 50 & 162 & 76.34 & 99.38 & 98.15 & 99.38 & \textbf{100.00} & \textbf{100.00} & \textbf{100.00} & 99.38 &
\textcolor{black}{\textbf{100.00}} & \textcolor{black}{98.76} & \textcolor{black}{\textbf{100.00}} & \textcolor{black}{\textbf{100.00}} & \textcolor{black}{\textbf{100.00}} \\
Woods & 50 & 1244 & 75.06 & 79.98 & 90.03 & 92.04 & 87.70 & 94.21 & \textbf{99.92} & 95.90 &
\textcolor{black}{98.95} & \textcolor{black}{99.59} & \textcolor{black}{92.44} & \textcolor{black}{97.67} & \textcolor{black}{99.12} \\
BGTD* & 50 & 330 & 73.78 & 67.58 & 81.52 & 89.39 & 93.33 & 89.40 & 98.18 & 97.58 &
\textcolor{black}{\textbf{99.69}} & \textcolor{black}{98.48} & \textcolor{black}{87.88} & \textcolor{black}{\textbf{100.00}} & \textcolor{black}{99.09} \\
Stone-Steel-Towers & 50 & 45 & 72.50 & \textbf{100.00} & \textbf{100.00} & \textbf{100.00} & \textbf{100.00} & 89.26 & \textbf{100.00} & \textbf{100.00} &
\textcolor{black}{\textbf{100.00}} & \textcolor{black}{\textbf{100.00}} & \textcolor{black}{\textbf{100.00}} & \textcolor{black}{\textbf{100.00}} & \textcolor{black}{\textbf{100.00}} \\
Alfalfa & 15 & 39 & 71.22 & 94.87 & 71.79 & 92.31 & \textbf{100.00} & 89.13 & 97.44 & 94.87 &
\textcolor{black}{\textbf{100.00}} & \textcolor{black}{\textbf{100.00}} & \textcolor{black}{84.62} & \textcolor{black}{\textbf{100.00}} & \textcolor{black}{94.87} \\
Grass Pasture Mowed & 15 & 11 & 69.94 & \textbf{100.00} & \textbf{100.00} & \textbf{100.00} & 90.91 & 88.99 & \textbf{100.00} & \textbf{100.00} &
\textcolor{black}{\textbf{100.00}} & \textcolor{black}{\textbf{100.00}} & \textcolor{black}{\textbf{100.00}} & \textcolor{black}{\textbf{100.00}} & \textcolor{black}{\textbf{100.00}} \\
Oats & 15 & 5 & 68.65 & \textbf{100.00} & \textbf{100.00} & \textbf{100.00} & \textbf{100.00} & 88.85 & \textbf{100.00} & \textbf{100.00} &
\textcolor{black}{\textbf{100.00}} & \textcolor{black}{\textbf{100.00}} & \textcolor{black}{\textbf{100.00}} & \textcolor{black}{\textbf{100.00}} & \textcolor{black}{\textbf{100.00}} \\
\hline
OA (\%) & & & 80.07 & 75.46 & 78.02 & 82.79 & 77.56 & 86.51 & 90.64 & 93.08 &
\textcolor{black}{\textbf{98.32}} & \textcolor{black}{97.06} & \textcolor{black}{81.67} & \textcolor{black}{97.92} & \textcolor{black}{96.04} \\
AA (\%) & & & 86.97 & 85.31 & 84.30 & 90.03 & 88.93 & 92.33 & 94.88 & 96.31 &
\textcolor{black}{\textbf{99.11}} & \textcolor{black}{98.39} & \textcolor{black}{88.77} & \textcolor{black}{98.88} & \textcolor{black}{97.61} \\
Kappa & & & 77.43 & 72.13 & 75.11 & 80.44 & 74.65 & 84.63 & 89.27 & 92.08 &
\textcolor{black}{\textbf{98.08}} & \textcolor{black}{96.64} & \textcolor{black}{79.18} & \textcolor{black}{97.61} & \textcolor{black}{95.46} \\
\hline
\end{tabular*}
\vspace{2pt}
\parbox{\textwidth}{\raggedright
\footnotesize \textit{* BGTD: Buildings-Grass-Trees-Drives class.}
}
\end{table*}

\setlength{\abovecaptionskip}{2pt}
\setlength{\belowcaptionskip}{2pt}

\setlength{\abovecaptionskip}{2pt}
\setlength{\belowcaptionskip}{2pt}
\begin{figure*}[t]
\centering
\includegraphics[width=\textwidth]{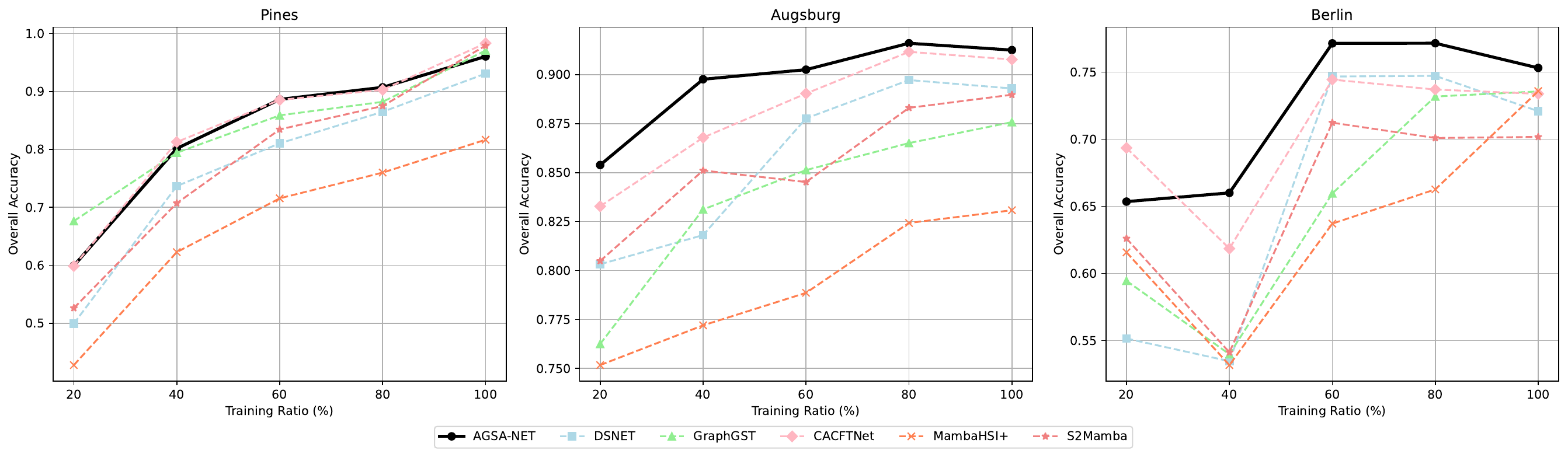}
\caption{\textcolor{black}{Effect of training sample ratio on overall accuracy (OA) across the evaluated datasets and different models, showing performance variations under different ratios of training data.}}
\label{fig:tsr_vs_oa}
\end{figure*}

\begin{table}[t]
\centering
\caption{Computational complexity comparison on the Indian Pines dataset (patch size $7\times7$). $\downarrow$ denotes that lower is better. \textbf{Bold} indicates the best result, and \underline{underlined} indicates the second-best result.}
\label{tab:complexity}
\setlength{\tabcolsep}{6pt}
\begin{tabular}{lccc}
\hline
\multirow{2}{*}{\textbf{Model}} & \textbf{\# of Params} & \textbf{FLOPs} & \textcolor{black}{\textbf{Inference Time}} \\
 & \textbf{(Millions)} $\downarrow$ & \textbf{(Millions)} $\downarrow$ & \textcolor{black}{\textbf{(ms)}} $\downarrow$ \\
\hline
DSNet~\cite{DCNET} & 0.42 & 11.00 & \textbf{0.021} \\
{\color{black}SSFTT \cite{SSFTT}} & {\color{black}\textbf{0.15}} & {\color{black}\textbf{1.43}} & {\color{black}\underline{0.022}} \\
{\color{black}GraphGST \cite{GraphGST}} & {\color{black}2.14} & {\color{black}\underline{2.15}} & {\color{black}0.094} \\
{\color{black}CACFTNet \cite{CACFTNet}} & {\color{black}1.86} & {\color{black}118.52} & {\color{black}0.285} \\
{\color{black}ViT \cite{ViT}} & {\color{black}5.37} & {\color{black}267.31} & {\color{black}0.265} \\
AGSA-Net (Ours) & \underline{0.24} & 11.48 & 0.045 \\
\hline
\end{tabular}
\end{table}

\subsection{Quantitative Analysis of Abundance Affinity}
\label{subsec:q_analysis_abu}
\rev{To examine whether the proposed abundance affinity matrix captures meaningful material relationships, we compare abundance-based similarity with raw spectral similarity. For each sample patch, the abundance map generated by the unmixing branch is normalized and spatially averaged to obtain a patch-level abundance vector. Similarity between two samples is then computed using the inner product of normalized abundance vectors, consistent with Eq.~\ref{eq:abundance_affinity}, while spectral similarity is computed using cosine similarity between the mean spectral signatures of each patch. As shown in Table~\ref{tab:similarity_comparison}, abundance representations consistently produce a larger similarity gap between same-class and different-class samples across all datasets, suggesting that abundance vectors encode more discriminative material relationships than raw spectral signatures and provide a stronger material-oriented prior for attention guidance.}

\rev{To further analyze whether the performance gain is associated with the specific guidance mechanism in Eq.~(4), we conduct an additional controlled intervention study, shown in Table~\ref{tab:guidance_intervention}. Unlike the ablation study in Table~\ref{tab:ablation}, where removing abundance-guided self-attention changes an entire architectural component, this experiment isolates the contribution of Eq.~(4) by removing abundance-feature fusion and allowing abundance information to affect classification only through the attention bias term. Four guidance strategies are evaluated: no guidance (standard self-attention), spectral guidance (spectral similarity prior), shuffled guidance (abundance affinity with disrupted material relationships), and the proposed abundance guidance. The results show a consistent trend across all datasets, where abundance guidance achieves the best performance, while spectral and shuffled guidance lead to reduced performance. In particular, shuffled guidance preserves the affinity statistics but destroys the material relationship structure, indicating that the improvement is associated with the material-aware structure encoded in the abundance affinity matrix rather than merely introducing an additional bias term.}

\begin{table}[t]
\centering
\small
\setlength{\tabcolsep}{4pt} 
\color{black}
\caption{Similarity Comparison Between Abundance and Spectral Representations}
\begin{tabular}{l l c c c}
\hline
\textbf{Dataset} & \textbf{Sim. Type} & \textbf{Same-Class} & \textbf{Different-Class} & \textbf{Gap} \\
\hline
Indian Pines & Spectral & 0.9939 & 0.9233 & 0.0706 \\
              & Abundance & \textbf{0.9562} & \textbf{0.8403} & \textbf{0.1159} \\
\hline
Augsburg & Spectral & 0.9769 & 0.9425 & 0.0344 \\
             & Abundance & \textbf{0.9671} & \textbf{0.8958} & \textbf{0.0713} \\
\hline
Berlin & Spectral & 0.9811 & 0.9535 & 0.0276 \\
             & Abundance & \textbf{0.8844} & \textbf{0.7857} & \textbf{0.0987} \\
\hline
\end{tabular}
\label{tab:similarity_comparison}
\end{table}

\begin{table*}[t]
\centering
\caption{Controlled intervention analysis of different attention guidance strategies.}
\label{tab:guidance_intervention}
\renewcommand{\arraystretch}{1.2}
\setlength{\tabcolsep}{5pt}
\footnotesize
\begin{tabular*}{\textwidth}{@{\extracolsep{\fill}}l|c|ccc|ccc|ccc}
\hline
\multirow{2}{*}{Description} &
\multirow{2}{*}{Equation} &
\multicolumn{3}{c|}{Indian Pines} &
\multicolumn{3}{c|}{Augsburg} &
\multicolumn{3}{c}{Berlin} \\
\cline{3-11}
& &
OA & AA & $\kappa$ &
OA & AA & $\kappa$ &
OA & AA & $\kappa$ \\
\hline
Shuffled Guidance &
$QK^\top+\lambda P A_{abu} P^\top$ &
80.72 & 88.40 & 78.01 &
84.32 & 62.47 & 77.33 &
68.92 & 62.97 & 55.47 \\
Spectral Guidance &
$QK^\top+\lambda A_{spec}$ &
79.73 & 88.32 & 77.03 &
84.68 & 63.27 & 77.87 &
70.65 & 64.44 & 57.30 \\
No Guidance &
$QK^\top$ &
80.92 & \textbf{89.55} & 78.35 &
84.96 & 60.66 & 78.24 &
72.00 & 60.54 & 57.69 \\
Proposed Abundance Guidance &
$QK^\top+\lambda A_{abu}$ &
\textbf{81.99} & 89.12 & \textbf{79.40} &
\textbf{85.57} & \textbf{64.00} & \textbf{79.25} &
\textbf{72.29} & \textbf{64.63} & \textbf{59.09} \\
\hline
\end{tabular*}
\end{table*}

\subsection{Impact of Training Sample Ratio}
\label{subsec:impact_tr}
We analyze the performance of AGSA-Net under varying levels of labeled data using class-wise training ratios of 20\%, 40\%, 60\%, 80\%, and 100\%, as illustrated in Fig.~\ref{fig:tsr_vs_oa}. In each setting, a fixed percentage of labeled samples per class is randomly selected for training using a stratified sampling strategy, while the test set remains unchanged. For example, in the Augsburg dataset, a class with 146 labeled samples at 100\% contributes approximately 29 samples at 20\%, 58 at 40\%, and 88 at 60\%, following a linear proportional reduction. The same sampling strategy is applied across all datasets, ensuring a fair comparison where performance differences are solely due to training data availability. As the training ratio decreases, all models experience performance degradation due to reduced supervision; however, AGSA-Net consistently achieves \rev{competitive performance across diverse datasets}. For instance, on Indian Pines at a 20\% training ratio, AGSA-Net achieves an overall accuracy of 0.5994, clearly outperforming DSNet at 0.4994 and MambaHSI+ at 0.4279, while also surpassing S2Mamba at 0.5263 and remaining competitive with GraphGST at 0.6764 and CACFTNet at 0.5987. Similar trends are observed on Augsburg and Berlin, where AGSA-Net maintains higher or comparable performance compared to other methods at low training ratios. Across intermediate ratios such as 40\% and 60\%, all models show steady improvement as more training samples become available, while AGSA-Net continues to maintain a consistent advantage or remains highly competitive. At higher ratios (80\% and 100\%), the performance gap between methods narrows, but AGSA-Net still achieves strong results without degradation. More importantly, AGSA-Net demonstrates a more stable performance trend across all ratios, with smaller accuracy reductions when moving from higher to lower training ratios compared to DSNet, MambaHSI+, and S2Mamba, which exhibit more pronounced declines. This indicates stronger robustness under limited labeled data conditions, making it suitable for real-world hyperspectral classification scenarios.

\subsection{Ablation on Component Contribution}
\label{subsec:ablation}

\rev{ 
An ablation study was performed to assess the contribution of AGSA-Net’s core components using OA, AA, and $\kappa$ as metrics (Table~\ref{tab:ablation}). The full AGSA-Net configuration achieves the strongest overall performance on Indian Pines and Augsburg, while maintaining competitive performance on Berlin, demonstrating the effectiveness of jointly integrating unmixing, abundance-guided self-attention, subpixel fusion, and spectral transformer reasoning. Removing abundance-guided self-attention (AGMHSA) leads to relatively modest performance changes, suggesting that subpixel fusion provides strong discriminative cues, with guidance mainly enhancing contextual stability. In contrast, disabling subpixel fusion causes a noticeable decline, highlighting the critical role of abundance feature fusion. Replacing the transformer with a CNN backbone (DSNet) gives intermediate performance, indicating that long-range spectral-spatial modeling via transformers is more effective. 
}

\rev{Although the impact of abundance-guided self-attention on overall accuracy (OA) is relatively modest, the ablation results indicate that it provides complementary contextual information within the proposed framework. Compared with abundance feature fusion, its contribution to OA is smaller, suggesting that AGMHSA primarily enhances contextual interactions rather than serving as the dominant source of discriminative information. 
In particular, on the Berlin dataset, removing abundance-guided self-attention results in slightly higher OA and $\kappa$ despite reduced performance on other datasets. This is attributed to the stronger contribution of abundance feature fusion in highly heterogeneous urban scenes, where it can partially compensate for the absence of attention guidance. This further supports that AGMHSA functions as a complementary contextual refinement mechanism rather than the primary source of discriminative power.
} 
Table \ref{tab:ablation} also presents the overall accuracy comparison of AGSA-Net, DSNet \cite{DCNET}, and the raw spectral Transformer across all three datasets. \rev{The results indicate that AGSA-Net outperforms the other methods in Indian Pines and Augsberg,} while using the Transformer alone yields the lowest performance. These ablation results show that each component of the proposed framework plays a complementary role in improving hyperspectral image classification performance, particularly in more complex urban environments.

\subsection{\rev{Visualization of Abundance-Guided Attention Behavior}}
\label{subsec:attention_visualization}
\begin{figure*}[t]
\centering
\includegraphics[width=\textwidth]{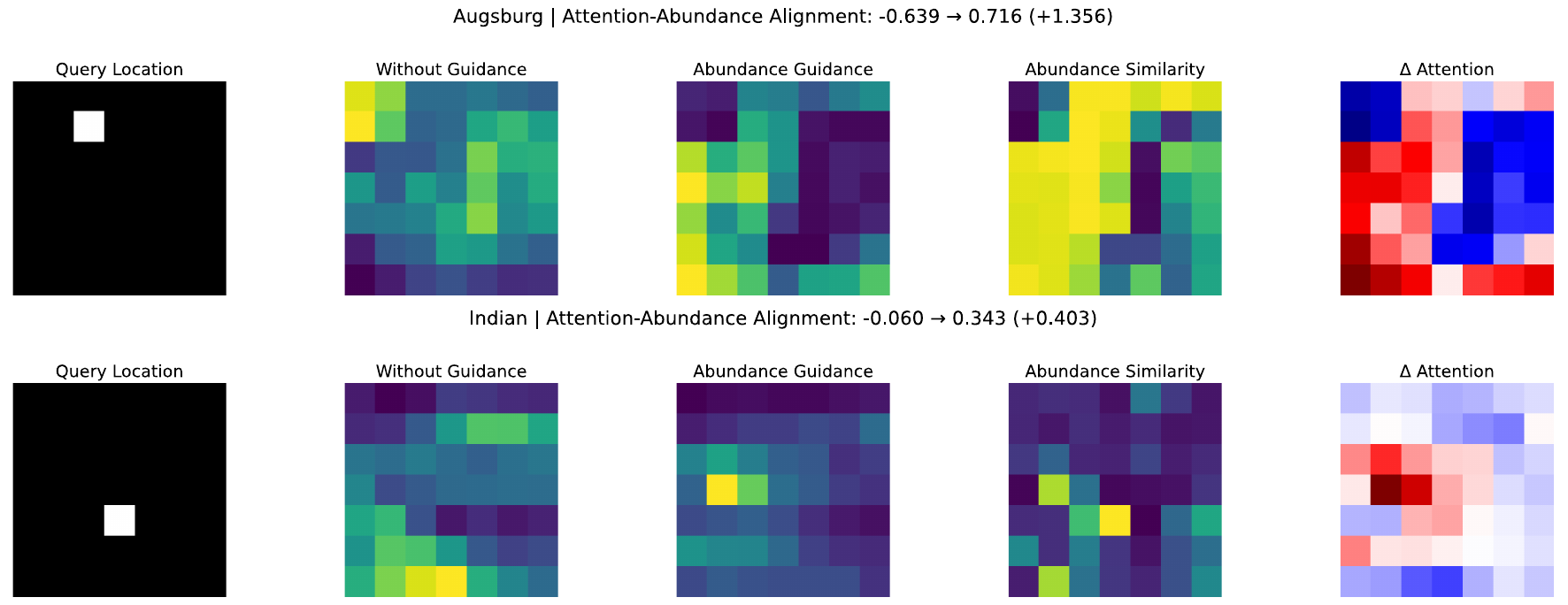}
\caption{Visualization of abundance-guided self-attention behavior for the Augsburg (top) and Indian Pines (bottom) datasets. For each dataset, the panels show: (1) query token location, (2) attention map without guidance, (3) attention map with abundance guidance, (4) abundance similarity map, and (5) attention difference ($\Delta$ Attention) between guided and unguided settings. The alignment score above each row denotes the correlation between the attention distribution and abundance similarity before and after introducing abundance guidance.}
\label{fig:attention_visualization}
\end{figure*}

\rev{To further analyze the effect of abundance-guided self-attention beyond classification accuracy, we examine how the attention distribution changes after introducing abundance affinity. For a given input sample, attention maps are extracted from the final transformer layer under two settings: (i) without abundance guidance and (ii) with abundance-guided attention enabled. A representative query token is selected automatically according to the largest increase in alignment between attention distribution and abundance similarity while excluding boundary tokens. Fig.~\ref{fig:attention_visualization} presents representative examples from the Augsburg and Indian Pines datasets. For each dataset, the panels show the query token location, attention distributions without and with guidance, abundance similarity, and the corresponding attention difference ($\Delta$ Attention). The alignment score above each row is computed as the correlation between the query-token attention distribution and abundance similarity, where higher values indicate stronger consistency between attention allocation and abundance-aware material relationships.}

\rev{As shown in Fig.~\ref{fig:attention_visualization}, abundance guidance redistributes attention toward regions with similar abundance characteristics. Augsburg exhibits a stronger restructuring effect, whereas Indian Pines shows a more moderate but consistent refinement behavior. These observations suggest that abundance-guided self-attention primarily acts as a contextual refinement mechanism that complements abundance feature fusion, consistent with Table~\ref{tab:ablation}, where abundance guidance provides complementary gains while abundance feature fusion contributes more directly to overall classification performance.}

\begin{table*}[t]
\centering
\caption{Core component analysis of the proposed AGSA-Net on Indian Pines, Augsburg, and Berlin datasets.}
\label{tab:ablation}
\renewcommand{\arraystretch}{1.2}
\setlength{\tabcolsep}{5pt}
\footnotesize
\begin{tabular*}{\textwidth}{@{\extracolsep{\fill}}cccc|c|ccc|ccc|ccc}
\hline
\multicolumn{4}{c|}{Modules} &
\multirow{2}{*}{Remarks} &
\multicolumn{3}{c|}{Indian Pines} &
\multicolumn{3}{c|}{Augsburg} &
\multicolumn{3}{c}{Berlin} \\
\cline{1-4}\cline{6-14}
UA & AGM & Fusion & Trans & & OA & AA & $\kappa$ & OA & AA & $\kappa$ & OA & AA & $\kappa$ \\
\hline
$\times$ & \rev{$\times$} & $\times$ & \checkmark & Transformer only
& 81.30 & 89.14 & 78.68
& 82.22 & 56.97 & 74.12
& 68.00 & 62.92 & 50.65 \\
\checkmark & \checkmark & $\times$ & \checkmark & No abundance fusion
& 81.99 & 89.12 & 79.40
& 85.57 & 64.00 & 79.25
& 72.29 & 64.63 & 59.09 \\
\checkmark & $\times$ & \checkmark & $\times$ & CNN backbone (DSNet)
& 93.08 & 96.31 & 92.08
& 89.30 & 67.20 & 84.62
& 72.09 & 65.34 & 59.27 \\
\checkmark & $\times$ & \checkmark & \checkmark & No abu-guided self-attn
& 94.72 & 97.36 & 93.96
& 90.48 & 67.41 & 86.26
& \textbf{76.04} & 64.77 & \textbf{63.50} \\
\checkmark & \checkmark & \checkmark & \checkmark & \textbf{AGSA-Net}
& \textbf{\textcolor{black}{96.04}} & \textbf{\textcolor{black}{97.61}} & \textbf{\textcolor{black}{95.45}}
& \textbf{91.26} & \textbf{68.47} & \textbf{87.46}
& 75.31 & \textbf{65.04} & 63.04 \\
\hline
\end{tabular*}
\end{table*}

\subsection{Analysis of the Abundance Guidance Weight $\lambda$}
\begin{figure}[t]
\centering
\includegraphics[width=\columnwidth]{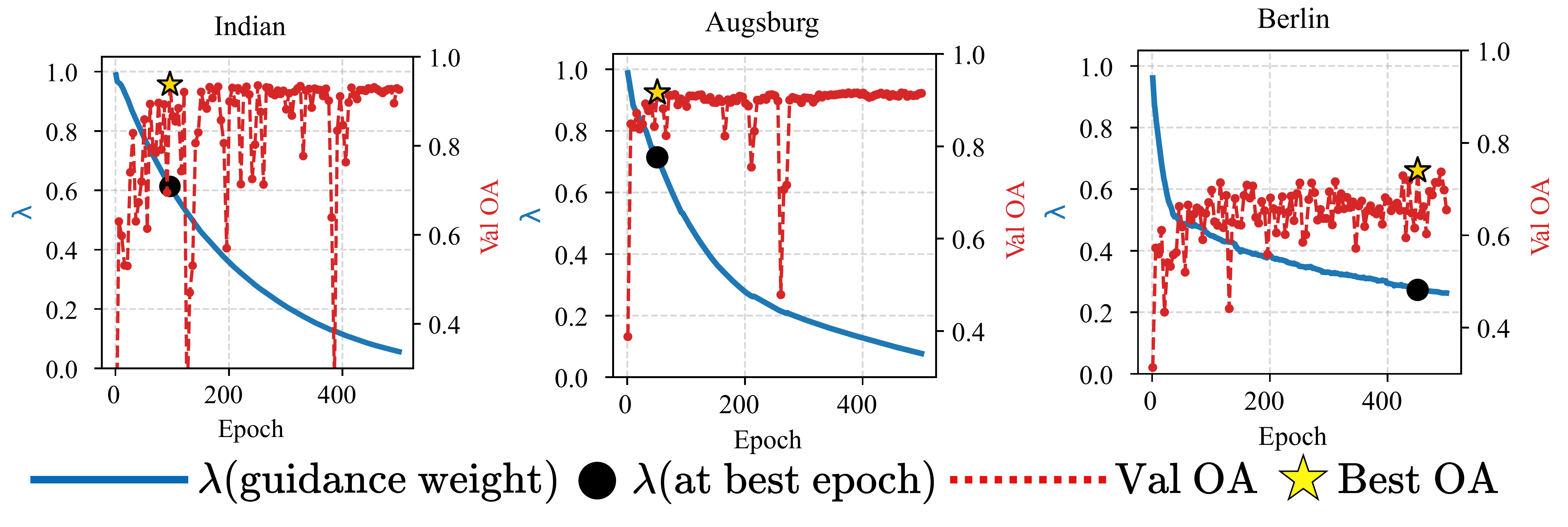}
\caption{Training dynamics of the abundance guidance weight $\lambda$ and validation accuracy on the Indian Pines, Augsburg, and Berlin datasets. The circle indicates the value of $\lambda$ at the epoch where the best validation accuracy is achieved, while the star denotes the highest validation accuracy during training.}
\label{fig:lambda_dynamics}
\end{figure}

The abundance-guided attention mechanism introduced in Eq.~\ref{eq:affinity_attention} includes a learnable scalar parameter $\lambda$ that controls the contribution of the abundance affinity matrix to the attention logits. This parameter determines how strongly material composition similarity influences contextual attention during training. To better understand the behavior of this parameter, we analyze its evolution during training together with the validation accuracy on the Indian Pines, Augsburg, and Berlin datasets. The parameter is initialized with a value of 1 and is optimized throughout the training process. During training, we record values of $\lambda$ at each epoch and visualize its dynamics along with validation overall accuracy (OA), as illustrated in Fig.~\ref{fig:lambda_dynamics}. We make several observations from this analysis. First, $\lambda$ decreases gradually during training across all datasets, indicating that the model adaptively adjusts the strength of abundance guidance as spectral–spatial representations become increasingly discriminative. Second, $\lambda$ remains strictly positive throughout training and does not collapse to zero, demonstrating that the abundance affinity consistently contributes to the attention mechanism rather than being ignored by the model. Third, the value of $\lambda$ at the epoch achieving the best validation accuracy remains moderate for all datasets, confirming that the optimal models still rely on abundance-guided contextual interactions. We also observe that the learned magnitude of $\lambda$ varies across datasets. For the relatively homogeneous agricultural scene in Indian Pines, and the semi-urban Augsburg dataset, $\lambda$ gradually decreases as spectral features become sufficient for discrimination. In contrast, for the more heterogeneous Berlin urban scene, $\lambda$ stabilizes at a higher level, suggesting that abundance guidance plays a more persistent role in capturing complex material relationships. These observations demonstrate that the proposed abundance-guided attention mechanism remains effective throughout training and that the learnable parameter $\lambda$ adapts automatically to the characteristics of each dataset. 

\subsection{\rev{Sensitivity Analysis of the Number of Endmembers ($K$)}}
\label{subsec:k_sens}

\rev{
The number of endmembers $K$ is an important design parameter in abundance-based hyperspectral representation learning. In hyperspectral unmixing, the number of endmembers generally does not correspond to the number of semantic classes, as a class may comprise multiple materials while different classes may share common materials. In this work, we adopt $K=U$, where $U$ is the number of semantic classes, following DSNet~\cite{DCNET} to obtain a compact latent abundance representation and maintain consistency with prior unmixing-based classification frameworks. Since the estimated abundances are used as structural priors for abundance-guided attention and feature fusion rather than explicit material identification, the objective is to learn a compact and discriminative latent representation rather than recover the true number of physical materials. To evaluate this design choice, we conduct a sensitivity analysis with $K \in \{0.5U,0.75U,U,1.25U,1.5U,2U\}$, as summarized in Table~\ref{tab:sensitivity_k}. AGSA-Net is relatively robust to moderate variations in $K$. The default setting $K=U$ achieves the highest OA and Kappa on Indian Pines and Berlin, while Augsburg performs slightly better at $K=1.25U$, likely due to its greater material heterogeneity and mixed-pixel characteristics. Performance degrades when $K$ is too small because the abundance representation becomes undercomplete, whereas excessively large values introduce redundancy and reduce representation compactness. Overall, these results indicate that $K=U$ provides an effective and compact default choice for AGSA-Net without implying a one-to-one correspondence between endmembers and semantic classes.
}

\begin{table}[t]
\centering
\caption{\rev{Sensitivity analysis of the number of endmembers ($K$). Best OA, AA and Kappa values for each dataset are highlighted in bold.}}
\label{tab:sensitivity_k}
\resizebox{\columnwidth}{!}{%
{
\begin{tabular}{llcccccc}
\hline
Dataset & Metric & 0.5U & 0.75U & U & 1.25U & 1.5U & 2U \\
\hline
\multirow{3}{*}{Pines} & Best OA & 0.9343 & 0.9413 & \textbf{0.9604} & 0.9431 & 0.9478 & 0.9455 \\
 & Best AA & 0.9645 & 0.9700 & \textbf{0.9761} & 0.9671 & 0.9694 & 0.9746 \\
 & Kappa & 0.9248 & 0.9328 & \textbf{0.9546} & 0.9348 & 0.9401 & 0.9376 \\
\hline
\multirow{3}{*}{Berlin} & Best OA & 0.7211 & 0.7187 & \textbf{0.7531} & 0.7465 & 0.7214 & 0.7503 \\
 & Best AA & 0.6501 & 0.6368 & 0.6504 & 0.6173 & \textbf{0.6693} & 0.5911 \\
 & Kappa & 0.5912 & 0.5847 & \textbf{0.6304} & 0.6170 & 0.5945 & 0.6131 \\
\hline
\multirow{3}{*}{Augsburg} & Best OA & 0.8889 & 0.9023 & 0.9126 & \textbf{0.9278} & 0.9143 & 0.9238 \\
 & Best AA & 0.6248 & 0.6215 & 0.6847 & \textbf{0.7027} & 0.6652 & 0.6934 \\
 & Kappa & 0.8392 & 0.8581 & 0.8746 & \textbf{0.8959} & 0.8758 & 0.8901 \\
\hline
\end{tabular}
}
}
\end{table}

\subsection{Feature Representation Analysis (t-SNE)}
\setlength{\abovecaptionskip}{2pt}
\setlength{\belowcaptionskip}{2pt}
\begin{figure}[t]
\centering
\scriptsize
\begin{minipage}{0.48\linewidth}
\centering
(a) Raw Spectral\\[2pt]
\includegraphics[width=\linewidth]{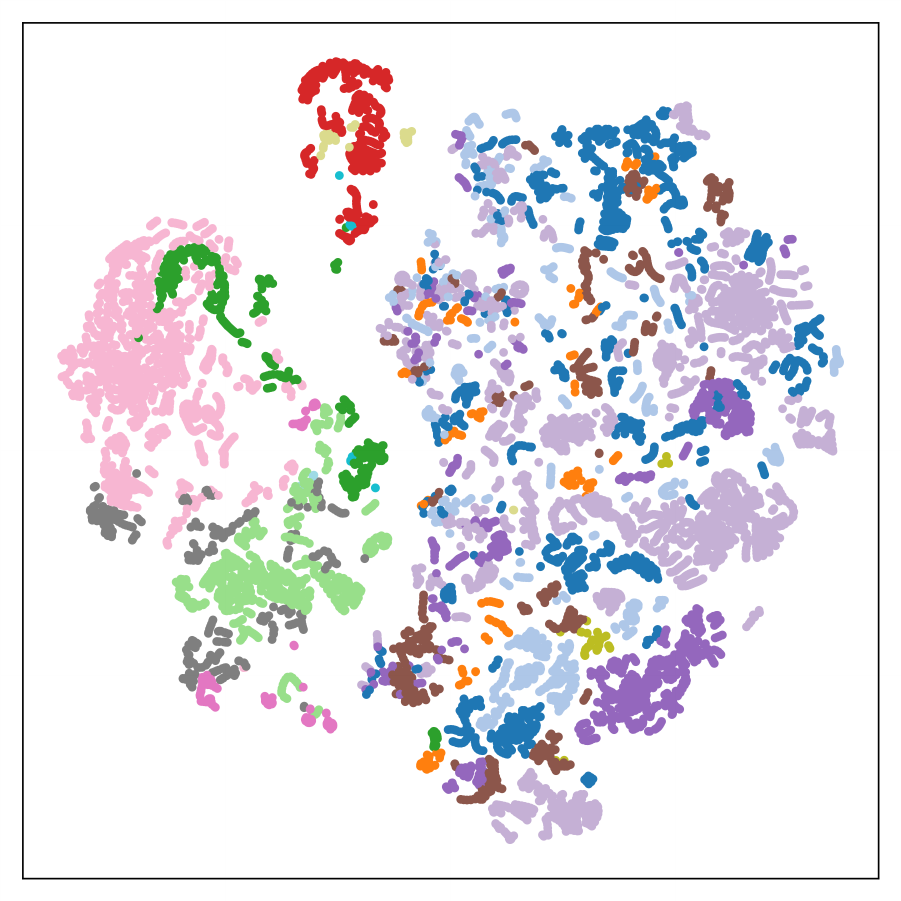}
\end{minipage}
\hfill
\begin{minipage}{0.48\linewidth}
\centering
(b) Transformer\\[2pt]
\includegraphics[width=\linewidth]{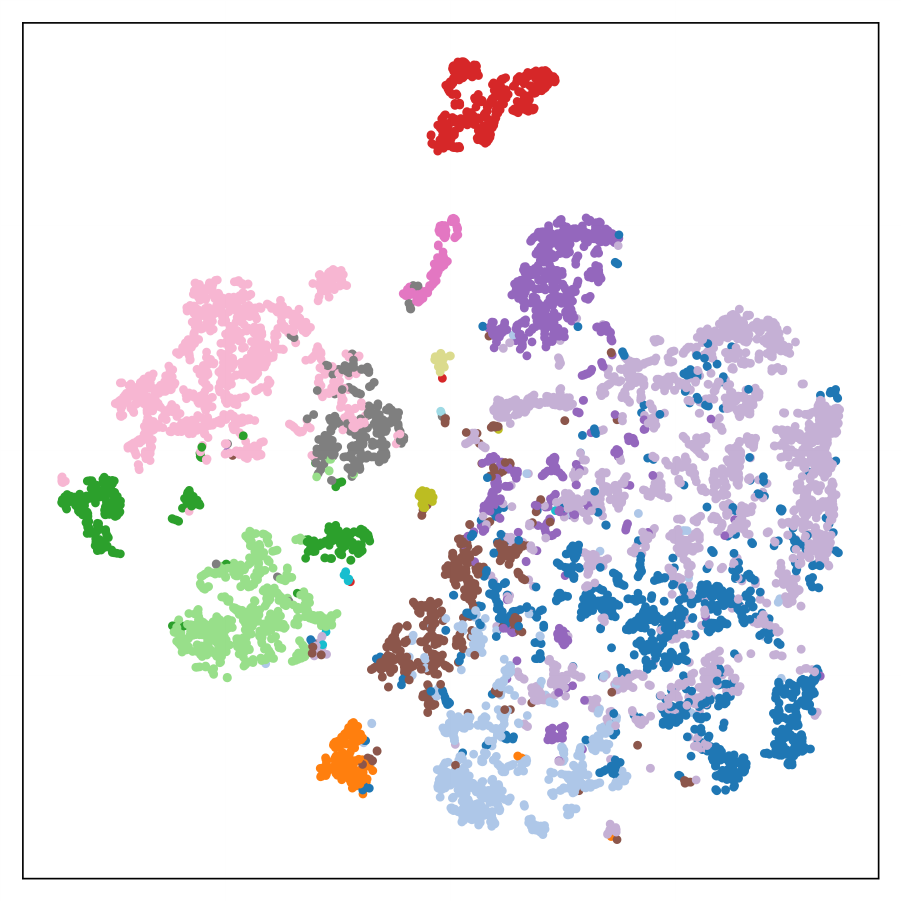}
\end{minipage}
\vspace{4pt} 
\begin{minipage}{0.48\linewidth}
\centering
(c) DSNet~\cite{DCNET}\\[2pt]
\includegraphics[width=\linewidth]{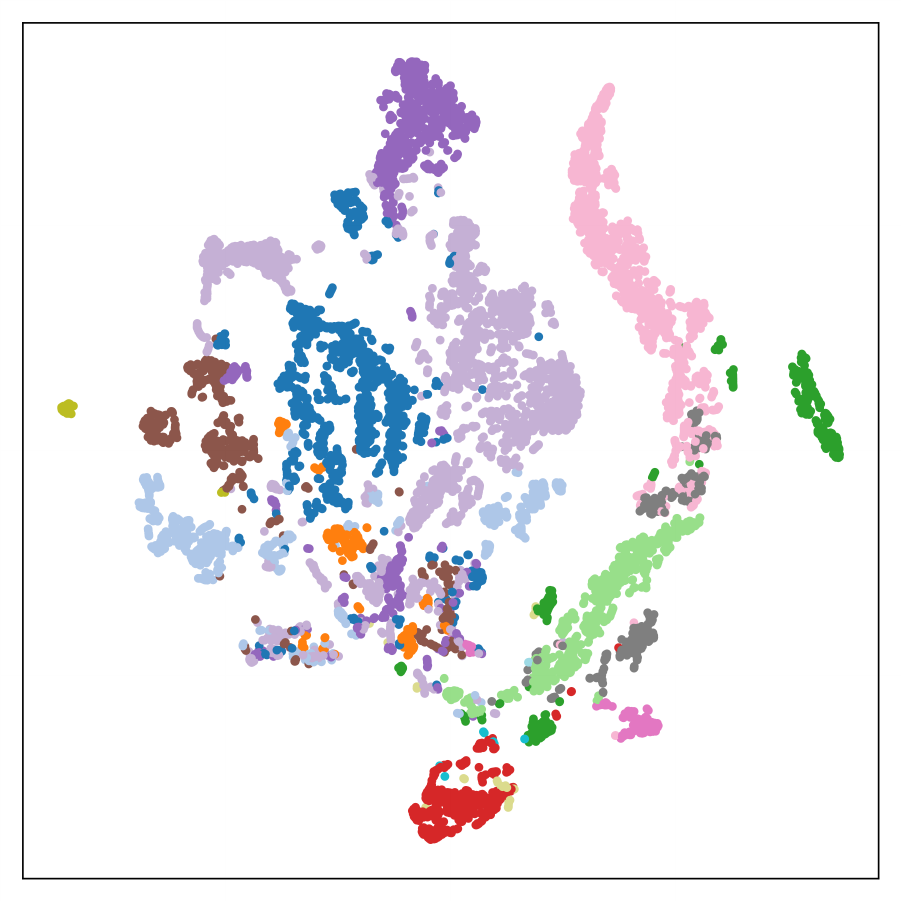}
\end{minipage}
\hfill
\begin{minipage}{0.48\linewidth}
\centering
(d) AGSA-Net (Ours)\\[2pt]
\includegraphics[width=\linewidth]{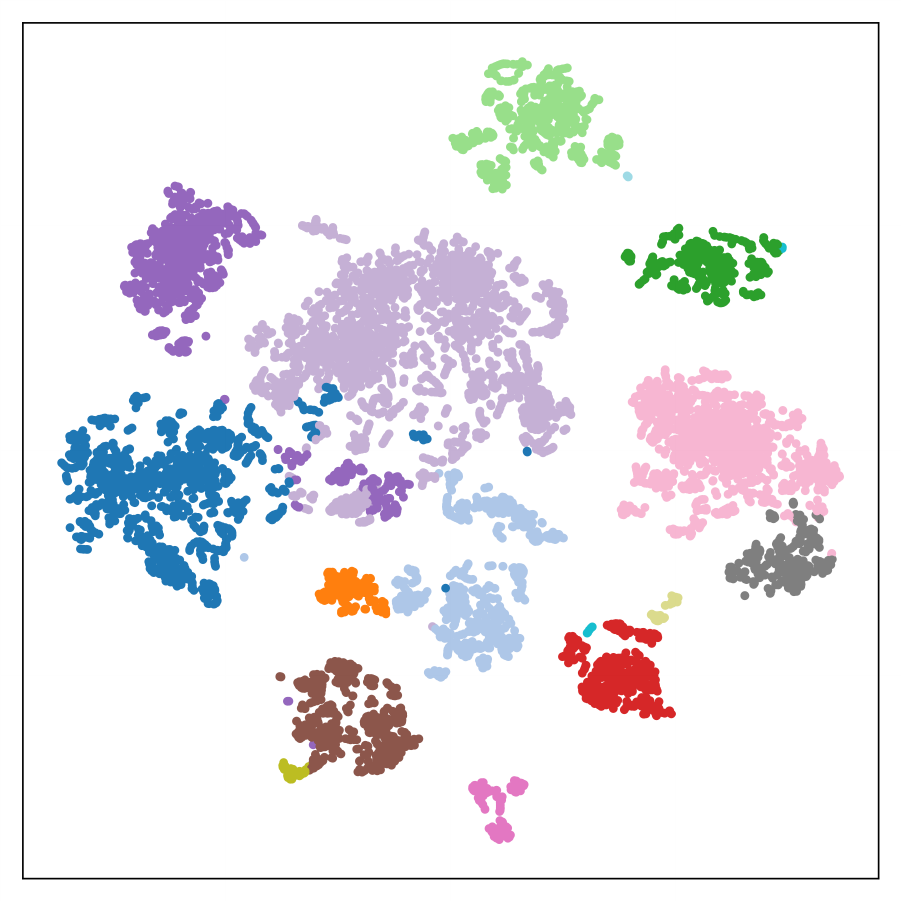}
\end{minipage}
\vspace{4pt}
\centerline{\includegraphics[width=\columnwidth]{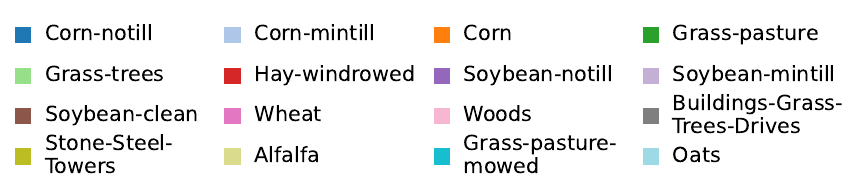}}
\caption{t-SNE visualization of feature representations from the Indian Pines dataset using (a) raw spectral features, (b) Transformer, (c) DSNet, and (d) AGSA-Net. AGSA-Net produces the most compact and well-separated clusters, indicating more discriminative feature representations.}
\label{fig:tsne}
\end{figure}
To further analyze the discriminative quality of learned features, t-SNE is used to visualize high-dimensional representations from the Indian Pines dataset (Fig.~\ref{fig:tsne}). Raw spectral features exhibit strong class overlap, reflecting limited separability among spectrally similar categories. Transformer-based features show partial improvement, but noticeable inter-class mixing remains. DSNet produces more compact clusters with improved separation, indicating the benefit of enhanced spectral-spatial modeling. In contrast, AGSA-Net yields the most distinct and well-separated clusters, with higher intra-class compactness and reduced overlap. These visual trends are consistent with the quantitative gains and confirm that AGSA-Net learns more discriminative and stable feature representations for HSI classification.

\section{Conclusion}
\label{sec:conclusion}
In this paper, we have introduced AGSA-Net, a dual-branch framework designed to improve hyperspectral remote sensing image classification. \rev{The proposed model integrates spectral unmixing with transformer-based classification by injecting abundance-derived material affinity directly into the self-attention mechanism. This enables the network to form contextual relationships based on sub-pixel material composition rather than spectral similarity alone, improving discrimination in mixed-pixel and heterogeneous scenes. Experimental results on multiple benchmark datasets demonstrate that AGSA-Net achieves competitive performance, particularly in complex urban environments, highlighting the value of physically informed attention for HSI classification.} 

\noindent\textbf{Limitations:}
Despite its effectiveness, AGSA-Net introduces additional computational cost due to transformer-based attention, and its performance depends on the quality of abundance maps, which may be affected by sensor noise. \rev{Future work will focus on more efficient and noise-robust abundance-guided attention, including sparse attention and compressed transformer designs, targeting a 2–3× reduction in FLOPs for real-time deployment on UAV-based edge platforms.}


\bibliographystyle{IEEEtran}
\bibliography{REF}

\end{document}